\documentclass[a4paper,fleqn]{cas-dc}

\usepackage[authoryear]{natbib}
\usepackage{soul}

\def\tsc#1{\csdef{#1}{\textsc{\lowercase{#1}}\xspace}}
\tsc{WGM}
\tsc{QE}
\begin{document}
\let\WriteBookmarks\relax
\def\floatpagepagefraction{1}
\def\textpagefraction{.001}

\shorttitle{Similarity learning for wells based on logging data}    

\shortauthors{E.Romanenkova et al.}  

\title [mode = title]{Similarity learning for wells based on logging data}  

%\tnotemark[] 

%\tnotetext[]{} 


\author[1]{Evgenia Romanenkova}[orcid=0000-0003-3623-0011]
\cormark[1]
\ead{shulgina@phystech.edu}
%\fnmark[<footnote mark no>]
%\ead[url]{<URL>}
\credit{Conceptualization, Methodology, Software, Validation, Investigation, Data Curation, Writing - Original Draft,  Writing - Review\&Editing, Visualization}

\author[1]{Alina Rogulina}
\credit{Methodology, Software, Validation, Investigation, Data Curation, Writing - Original Draft, Writing - Review\&Editing, Visualization}

\author[2]{Anuar Shakirov}
\credit{Formal analysis, Data Curation, Writing - Original Draft,  Writing - Review\&Editing, Visualization}

\author[1]{Nikolay Stulov}
\credit{Methodology, Software, Validation, Investigation,  Writing - Original Draft,  Writing - Review\&Editing, Visualization}

\author[1]{Alexey Zaytsev}[orcid=0000-0002-1653-0204]
\credit{Conceptualization, Writing - Original Draft,  Writing - Review\&Editing, Supervision, Project administration}

\author[2]{Leyla Ismailova}
\credit{Conceptualization, Methodology, Writing - Original Draft,  Writing - Review\&Editing, Project administration}

\author[2]{Dmitry Kovalev}
\credit{Conceptualization, Methodology, Writing - Original Draft,  Writing - Review\&Editing, Validation}

\author[3]{Klemens Katterbauer}
\credit{Conceptualization, Writing - Original Draft,  Writing - Review\&Editing}

\author[3]{Abdallah AlShehri}
\credit{Conceptualization, Writing - Original Draft,  Writing - Review\&Editing}

\affiliation[1]{organization={Skolkovo Institute of Science and Technology},
            %addressline={Bolshoy Boulevard 30, bld. 1}, 
            city={Moscow},
            %postcode={121205}, 
            %state={},
            country={Russia}}
\affiliation[2]{organization={Aramco Moscow Research Center, Aramco Innovations},
            %addressline={Leninskiye Gory 1-75B}, 
            city={Moscow},
            %postcode={119234}, 
            country={Russia}}
\affiliation[3]{organization={Saudi Aramco},
            city={Dhahran},
            country={Saudi Arabia}}

\cortext[1]{Corresponding author at Skolkovo Institute of Science and Technology, Moscow, Russia}



\begin{abstract}
One of the first steps during the investigation of geological objects is the interwell correlation. It provides information on the structure of the objects under study, as it comprises the framework for constructing geological models and assessing hydrocarbon reserves. Today, the detailed interwell correlation relies on manual analysis of well-logging data. Thus, it is time-consuming and of a subjective nature. The essence of the interwell correlation constitutes an assessment of the similarities between geological profiles. There were many attempts to automate the process of interwell correlation by means of rule-based approaches, classic machine learning approaches, and deep learning approaches in the past. However, most approaches are of limited usage and inherent subjectivity of experts. 
We propose a novel framework to solve the geological profile similarity estimation based on a deep learning model. 
Our similarity model takes well-logging data as input and provides the similarity of wells as output. The developed framework enables (1) extracting patterns and essential characteristics of geological profiles within the wells and (2) model training following the unsupervised paradigm without the need for manual analysis and interpretation of well-logging data. 
For model testing, we used two open datasets originating in New Zealand and Norway. Our data-based similarity models provide high performance: the accuracy of our model is $0.926$ compared to $0.787$ for baselines based on the popular gradient boosting approach.
With them, an oil\&gas practitioner can improve interwell correlation quality and reduce operation time.
\end{abstract}


% TODO

%\item We propose a data-based model that can calculate the similarity between well intervals. The model has an architecture based on a recurrent neural network and a  training scheme based on ideas from unsupervised and self-supervised learning.
%\item We develop a workflow that allows the use of raw data from a well to produce well similarities between interval similarities calculated via a data-based model proposed above. It is of separate interest and can be applied for the processing of sequential data for other oil\&gas problems.
%\item We present a model evaluation protocol that looks at the proposed similarity model from different points of view crucial for the well similarity model usage.
%\item For all steps, we consider different design choices and argument why a particular decision was made. 

% Keywords

\maketitle

% Main text
\section{Introduction}\label{sec:intro}

The lithological and physical properties of rocks can vary laterally across the basin within one formation due to changes in sedimentation conditions. It affects the logging signals and complicates the interwell correlation. Manual comparison of geological profiles and identification of similar ones can be a challenging problem. At the same time, reliable interwell correlation is the key component for constructing a geological model of a field and estimating hydrocarbon reserves. To date, geologists and petrophysisists, in most cases, process and interpret logging data manually. Concurrently, manual processing of logging data during interwell correlation is a time-consuming and expensive process. Moreover,  the result of processing might have a subjective nature. That means that the same data can be interpreted by another expert differently.

Recent solutions in the oil\&gas industry emphasize importance of data-based solutions that can reduce the amount of manual labour and are less prone to errors~\cite{xu2019petrophysics,tekic2019disruptively,koroteev2021artificial}. 

In the oil\&gas industry, it is common to use complex multi-component systems with a small added value of data mining~(\cite{murtazin}). 
A typical approach is to apply a data-based model on top of a carefully prepared dataset with a constant guidance from area experts.
Some works try to challenge this framework while still using classic machine learning approaches~\cite{gurina2020application, brazell2019machine}.
In other industries, the leaders aim at end-to-end deep neural networks that extract knowledge automatically from data in, e.g. natural language processing~\cite{devlin2018bert}, financial industry~\cite{babaev2020event} and earthquake prediction~\cite{kail2020recurrent}. An end-to-end deep neural network extracts meaningful representations from data and  applies them for similarity identification and solution of a particular downstream task, e.g. predict credit score for a bank client or fraud in medical insurance by comparing representations of different objects~\cite{snorovikhina2020unsupervised}.

Moreover, while the amount of labelled data is limited, deep neural networks open an opportunity for adaptation of unsupervised learning that doesn't need such data and can learn from raw data, which comprises most of the data in the industry nowadays.
Deep learning unsupervised methods leverage only unlabelled data to learn complex data representations~\cite{jing2020self,deng2019arcface}.
Typically, one applies a self-supervised approach for unsupervised learning when we try to learn data representations based on a natural similarity between objects.
For example, if intervals belong to the same well and are close in depth, their representations should be similar, and they should be different otherwise.

Taking into account the nature of logging data, it is reasonable to construct a similarity learning model for oil\&gas wells via deep neural networks. For similarity learning, we plan to use unlabelled data that is most often available in applied settings. The main idea is to learn a model that distinguishes if two objects come from a similar source or different ones~\cite{jaiswal2021survey}.
This approach can, e.g. include augmentation of an initial data piece and comparing it to the initial version.
For a pair of objects (e.g. well or well intervals), the learned model outputs the distance model between a pair of complex objects. Both deep learning and machine learning approaches can solve the similarity learning problem.

The article structure is the following:
\begin{itemize}
    \item In Section~\ref{sec:innovation} we make claims about innovations presented further in our work.
    \item In Section~\ref{sec:related} we describe the current state-of-the-art on similarity learning and its' application to oil\&gas data.
    \item In Section~\ref{sec:data} we provide a description of the used data.
    \item In Section~\ref{sec:methods} we give the formal problem definition and our approach to solve it. Then we describe used models and used validation schemes in detail. 
    \item In Section~\ref{sec:results} we provide the results of numerical experiments to benchmark our model.
    \item We end up with conclusions in Section~\ref{sec:conclusions}.
\end{itemize}

\section{Main contributions}\label{sec:innovation}

\paragraph{Unsupervised learning for similarity model training.}

\begin{figure*}[!ht]
    \centering
    \includegraphics[width=0.7\linewidth]{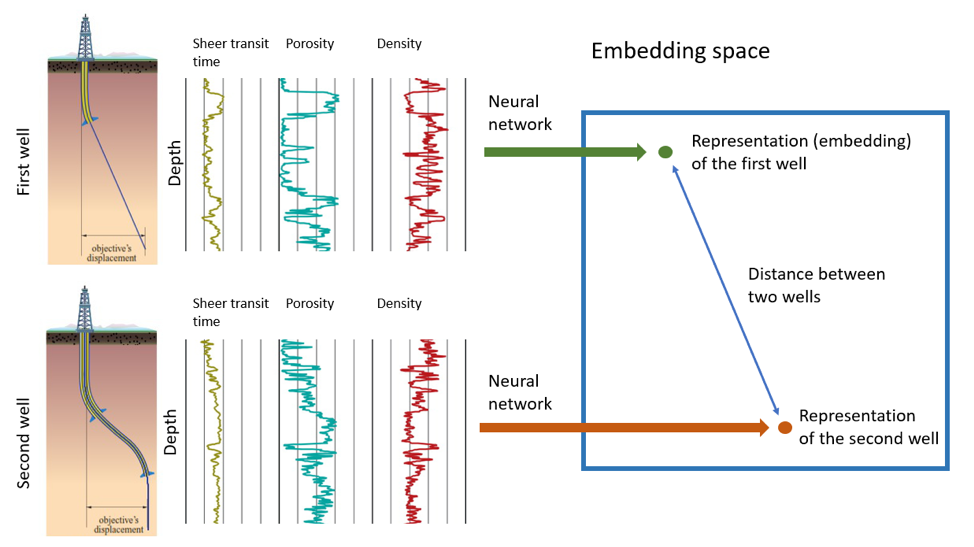}
    \caption{Principal scheme for the calculation of data-based distance model between wells. We collect data from two wells, then a neural network produces a representation of each well on base on logging data. Representations aggregate main information about wells and allow similarity evaluation.}
    \label{fig:tt-one}
\end{figure*}

The principal scheme of the work of our model at well-level is given in~\ref{fig:tt-one}.
We apply the similarity learning paradigm to the training of the similarity model based on well-logging data.
Our similarity model has two parts that follow each other: an encoder and a similarity evaluation (see~\ref{sec:methods}). The first encoder part produces data representations from logging data that characterise a well. 
The second comparison part estimates similarities between such intervals using obtained representations.
The comparison part can be simple and even be a single formula like Euclidean distance between embeddings.

We train both parts simultaneously in a self-supervised manner.
The training aims at making close representations for similar intervals from a single well, while representations of distinct intervals (in terms of lithology and petrophysical properties) from the same or other wells to be dissimilar.
As we know a well label and depth for each interval, we can train this model without expert labelling. 
This approach is only an approximation of the real world, as it is not always the case that intervals from different wells have different properties also intervals from the same well can be diverse.
However, it provides a good starting point and uses realistic assumptions. 
We prove that it provides reasonable results that are good enough for the usage of such an approach in practice. 

The self-supervised paradigm~\cite{oord2018representation,schroff2015facenet} requires careful selection of methods, data preparation, accounting for data peculiarities during model training~\cite{deng2019arcface}.
We compare different representation learning approaches and identify their strength and weakness for a particular problem of similarity learning for well-logging data.
Obtained models are evaluated against classic machine learning baselines  to answer two questions: 1) do our machine learning models for similarity estimation solve the particular problem at hand; 2) are obtained similarities meaningful from an expert point of view.

\begin{figure*}[!ht]
    \centering
    \includegraphics[width=0.8\textwidth]{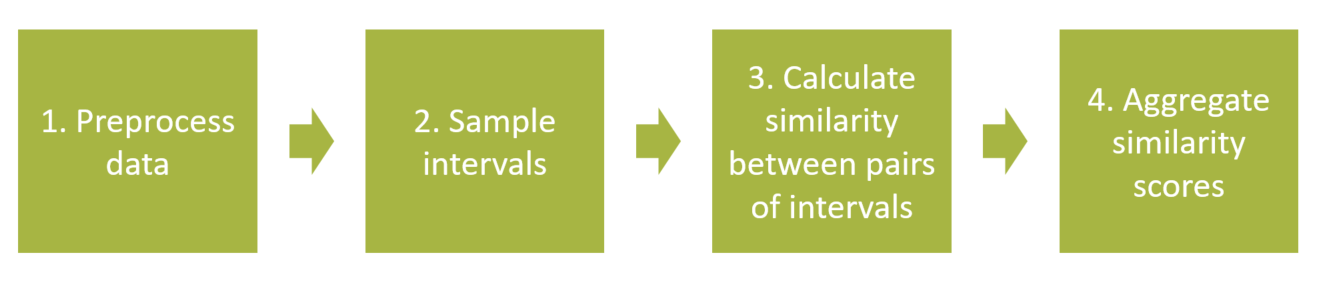}
    \caption{Workflow of similarity learning model application that allows model learning without labels provided by experts}
    \label{fig:workflow} 
\end{figure*}

\paragraph{Framework for evaluation of similarity between wells.}
The scheme above leads to a proposed framework for the application of the proposed similarity model.
It is depicted in Figure~\ref{fig:workflow}, that consists of four vital steps:
\begin{enumerate}
    \item \emph{Preprocessing:} transforms initial raw data and gets it ready to be used as an input to the encoder. It is of great importance to keep input format for diverse data similar and exclude unrealistic measurement and proceed with missing data.
    \item \emph{Sampling:} a well contains too distinct intervals, so it makes sense to consider intervals from a well and get similarities for them. Sampling of intervals allows usage of similarity model trained on pairs of intervals. Also, we can't train this model using the whole dataset.
    \item \emph{Similarity calculation:} get similarity scores between pairs of intervals using our data-based model.
    \item \emph{Aggregate similarity scores:} we get back from the level of intervals to the level of wells by aggregating scores obtained during the previous calculation step.
\end{enumerate}

\paragraph{Similarity model evaluation.}
We propose an evaluation procedure that can compare different obtained representations to make reasonable decisions on the best self-supervised approach.
To check the limitations of our approach, we do an ablation study of our approach to identify the components that affect the quality most.
A separate line of research considers the possibility of trained models to data from other fields never seen during training and how good it is for transfer learning.
We also examine how data labelling data can help with improving our models.

\paragraph{Summary.}

We present a first neural network-based model for interval similarity evaluation based on logging data. 
The novelty of the research is supported by the following: 
\begin{itemize}
    \item We propose a data-based model that can calculate the similarity between well intervals. The model has an architecture based on a recurrent neural network and a  training scheme based on ideas from unsupervised and self-supervised learning.
    \item We develop a workflow that allows the use of raw data from a well to produce well similarities between interval similarities calculated via a data-based model proposed above. It is of separate interest and can be applied for the processing of sequential data for other oil\&gas problems.
    \item We present a model evaluation protocol that looks at the proposed similarity model from different points of view crucial for the well similarity model usage.
    \item For all steps, we consider different design choices and argument why a particular decision was made. 
\end{itemize}

\section{Related work}\label{sec:related}

In this section, we consider current developments on similarity learning related to the oil\&gas industry.
We focus on data-based approaches as the quality and amount of available data increase, and there is an emerging trend in the usage of such models in our industry.

Identifying similarities between wells in the oil\&gas industry is a necessary part of such procedures as understanding the well's characteristics, planning field development strategy, detecting anomalies during the drilling process, generation of synthetic data closed to the real one, and many other applications \cite{verma2014assessment, gurina2020application, akkurt2018accelerating}. The current methods for estimating well-to-well correlation mainly concentrate on almost manually analysing features by experts or evaluating some non-natural similarity coefficient, e.g., simple Euclidean distance or more complex synchronization likelihood \cite{zoraster2004curve, verma2014assessment}. These classical approaches have in common low complexity of models and, consequently, a set of restrictions for their application. They often stay on the shoulders of high-qualified experts. In particular, they require a smart selection of inputs' features and model's hyperparameters. At the same time, modern machine learning and deep learning techniques have gained popularity in the oil\&gas industry, and we see that we can adapt such powerful paradigms for similarity learning \cite{gurina2020application, brazell2019machine}. These methods automatically extract features or representations and, thus, work without experts' specific subjective knowledge.
Below we expand these ideas by introducing simple rule-based approaches, classic machine learning approaches and deep learning approaches.

\emph{A rule-based approach} by~\cite{startzman1987rule} proposes to identify logical rules for the well-to-well correlation and identification of zones in wells' intervals based on prior experts knowledge. 
~\cite{jong1999interwell} move forward and show that using linear data-based model principal component analysis (PCA) combined with hierarchical cluster analysis performs better than the previous rule-based model.  
~\cite{zoraster2004curve} consider alignment by dynamic time warping (DTW) based on typical geometrical distances. After the curves, in particular, the authors mentioned Gamma-Ray curves, are aligned, the expert decides if they are similar. 
The paper~\cite{verma2014assessment} proposed a method for estimation similarity based on statistical methods. 
The authors calculate synchronization likelihood and visibility graph similarity to estimate closeness for intervals from original Gamma-Ray curves.
Although mentioned approaches are pretty simple for realization and easy-interpret, they primarily concentrate on considering one particular property from well-log data losing valuable information. In addition, most of them demand expert's involvement.

In recent years, \emph{classic machine learning} \cite{akkurt2018accelerating, gurina2020application} and deep learning \cite{ali2021machine, brazell2019machine} approaches have achieved more success in the development of geological models. In two papers by ~\cite{akkurt2018accelerating, ali2021machine}, authors consider similarity learning as one stage of their pipeline for the log-reconstruction system. They use Support Vector Machine (SVM) for obtaining smooth data’s representation named footprint and then apply common clusterization distances as Jaccard and Overlap distances to calculate the similarity between wells’ footprints. The most similar wells are used for predicting desired log with Quantile Regression Forest ~\cite{akkurt2018accelerating} and Fully Connected Neural network ~\cite{ali2021machine}. 
%Although the method shows reasonable results from experts’ point of view, such distances don’t consider the data’s nature, rely on carefully selected features, and can’t identify important representations of a well. 
Although the method shows reasonable results confirmed by experts’ evaluation, such distance model ignores data’s nature, while relying on carefully selected features. Thus, the method can’t identify representations of a well and is of limited applicability tailored to the considered ranges of selected features.
~\cite{gurina2020application} introduce similarity learning based on Measurement While Drilling (MWD) data for accident detection during drilling. The authors demonstrate the adoption of supervised similarity learning with expert-generated features: they aggregate statistics for time intervals and learn a classifier that predicts if intervals correspond to different or similar accidents. The obvious restriction is in manual obtaining labelled datasets for training such a system, as we need hundreds of labelled accidents.

Via \emph{deep learning}, artificial neural networks perform correlation estimation for wells. These data-based models use different features as input: oil production, water cut, injection pressure ~\cite{du2020connectivity}, permeability characterstics ~\cite{cheng2020lstm} or LAS-files \cite{brazell2019machine}.
~\cite{du2020connectivity} utilizes Convolution Neural Network for predicting regional permeability based on a set of geological features. Then, they proposed to use the average permeability of different regions calculated by the deep learning model as a similarity coefficient. The paper by~\cite{cheng2020lstm} raises an important question of estimation correlation between production and injection wells. The authors use Long Short-Term Memory Network for predicting the relationship between production wells and surrounding injection wells based on liquid production and water injection. After training the network, they apply techniques from sensitivity analysis to obtain inter-well connectivity coefficients. The work by~\cite{brazell2019machine} suggests using modern Convolution Neural Network for similarity evaluation between wells' LAS-images. To train the network, the authors consider images from a large dataset US Lower 48. Although these works are close to our approach, there are several important differences: the authors consider other input features and problem-statement or based on expert's labelling for training the model. At the same time, we propose an unsupervised approach for pure well-logs. Nevertheless, they show that produced similarities are highly correlated with the experts' ones. 

Recent deep metric learning advances~\cite{kaya2019deep} highlights its success in numerous applications in different industries. These approaches base their decision only on data and learn representations of structured data, including time series. 
For the oil\&gas industry, the list of possible similarity learning applications is long, while the amount of area-specific existing methods is downhearted. Most approaches are of limited usage and the inherent subjectivity of experts. 
We expect to fill this gap. 
This work introduces an approach for similarity learning based on well-log data in an unsupervised manner that overcomes mentioned restrictions and performs well from both expert's and machine learning's points of view.

\section{Data overview and preprocessing}\label{sec:data}

We utilized two datasets from New Zealand's Taranaki Basin and Norway's offshore for the task at hand. 
The larger dataset from Taranaki Basin was used for model training and validation, whereas the smaller dataset from Norway was used to assess the generalization ability of the developed models. Both datasets are open access.
The considered well-logging data are representative, large-scale and suitable for the similarity learning problem.
They share the same logging data while representing distinct basins located far away from each other. Therefore, we can reliably assess the generalization abilities of the similarity learning models. 

\subsection{Data details} 

\begin{figure*}[!ht]
    \centering
    \includegraphics[width=0.8\linewidth]{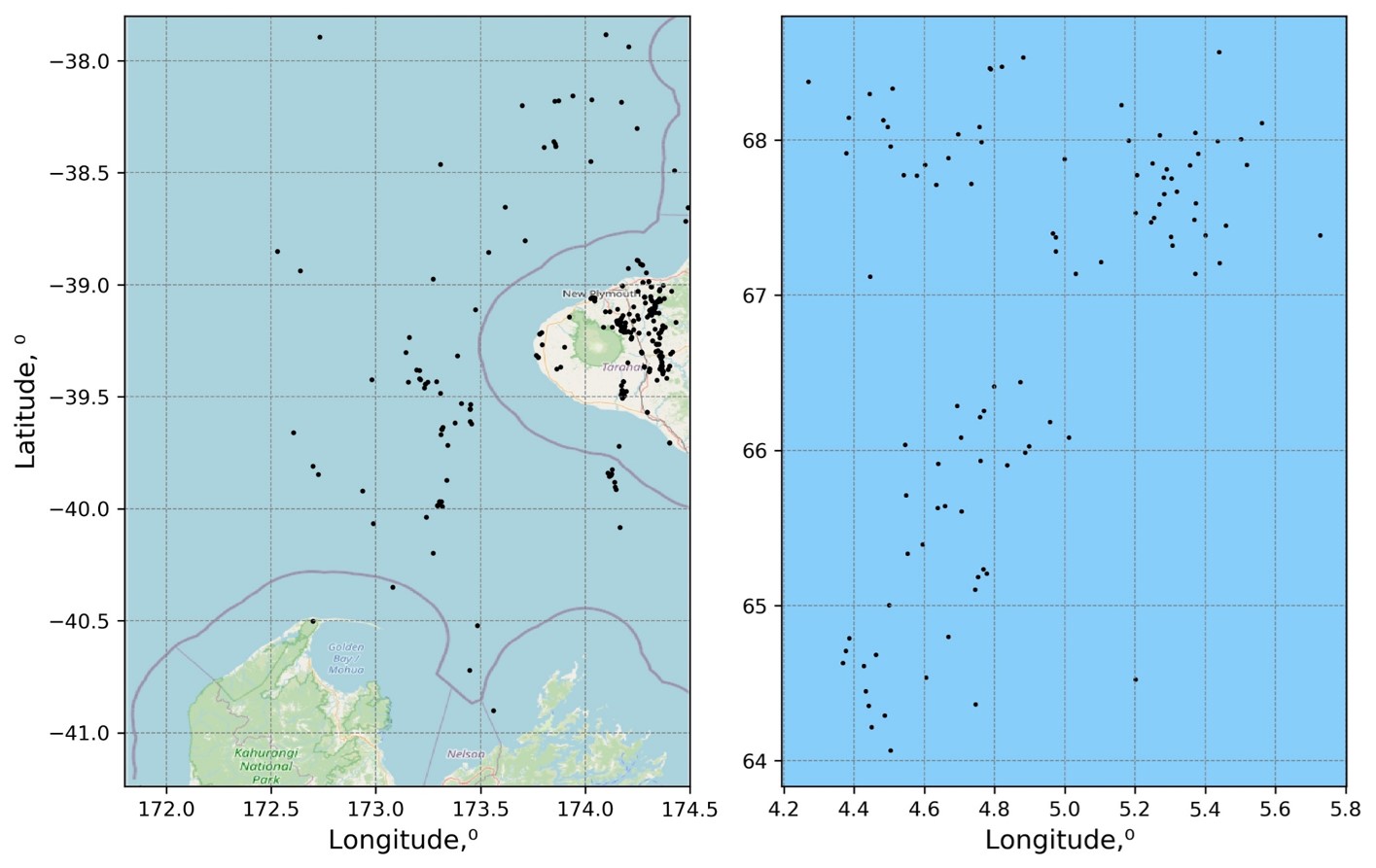}
    \caption{Location of wells from the Taranaki Basin (left) and Norway's offshore (right) available in the datasets. Black dots stand for the wells location}
    \label{fig:coord}
\end{figure*}

The data from Taranaki Basin was provided by the New Zealand Petroleum \& Minerals Online Exploration Database \citep{web:newzeland1}, and the Petlab~\citep{web:newzeland2}. Norwegian Petroleum Directorate~\citep{web:norwayweb} yield the open dataset from the Norwegian offshore. Figure \ref{fig:coord} shows locations of wells available in the Taranaki Basin and Norway datasets.

The used dataset from the Taranaki Basin includes the following $20$ well-logs: calliper (CALI), bit size (BS), gamma-ray (GR), neutron log (NEUT), density log (DENS), spontaneous potential log (SP), photoelectric factor log (PEF), sonic log (DTC), porosity inferred from density log (DRHO), deep resistivity (RESD), medium resistivity (RESM), shallow resistivity (RESS), tension (TENS), temperature log (TEMP), measured depth (DEPT), X, Y, Z well coordinates, well and formation name for each point. The dataset from Norway includes the same logging suite. The available logging suite is quite often standard within various geophysical companies. Therefore, the obtained results and models can be transferred to other basins. 

The dataset from Taranaki Basin includes 21 unique geological formations. During our research, we used data from five formations presented in Table \ref{tab:formations}. The dataset from Norway includes 68 geological formations, and for validation, we used only data from the Utsira formation presented in the same table.
The formations are comprised of both terrigenous (sandstone, siltstone, mudstone etc.) and carbonate rocks (limestones and dolomite).

\begin{table*}[h!]
\centering
\caption{Geological formations presented in the dataset and the number of wells drilled through each formation.}
\begin{tabular}{lcccc}
\hline
Formation & Number & Measurement  & Facies & Porosity,\%\\ 
& of wells & points & &\\ \hline
Urenui &219 & 848895 & slope sandstones & $\geq 15$ \\
Manganui &156 & 650398 & fine-grained basinal clactic rocks & $0-20$ \\
Matemateaonga & 148 & 613641 & sandy mudstones and siltstones & $\geq 15$ \\
Mohakatino & 86 & 516179 & volcaniclastic and epiclastic deposits & $5-15$ \\ 
Moki &76 & 265690 & deepwater turbidite sandstones & $14-20$ \\
Utsira & 40 & 172636 & sandstones & $\geq 30$ \\
\hline
\end{tabular}
\label{tab:formations}
\end{table*}

Since each formation was formed in unique sedimentation conditions, it resulted in (1) distinct lithological composition, (2) different petrophysical characteristics of rocks (even, in some cases, for one lithological type from different formations) and (3) distinct logging signals. Therefore, during similarity learning model training, the knowledge on geological profile stratigraphy is required to indicate the difference between considering depth intervals or wells from a geological point of view. 

\subsection{Expert's labelling}\label{sec:data_labeling}

The qualitative and quantitative analysis of well-logging data revealed that essentially different types of geological profiles occur within the same stratigraphic units. It is due to lateral changes in the sedimentation environment within the  basins. To take this peculiarity into account during the models' development and evaluation, we completed the expert labelling of wells based on the following similarity criteria:

\begin{enumerate} 
 \item The structure of the geological profiles is similar. For the structure of geological profiles, we imply the pattern of layer alternation.
 \item Lithological composition of wells is similar.
 \item Interrelations between rock physical properties (inferred from well-logging) are similar.
\end{enumerate}

We analysed geological profiles of wells within formations presented in Table \ref{tab:formations} and assigned the classes to all wells. The total number of geological profile classes are summarized in Table \ref{tab:labeling}. Additionally, within each geological profile class, we conducted interwell correlation to trace similar layers. The geological profile classes within each formation are unique. It means that the assigned digit labels (indicating profile classes) for wells within one formation do not correspond to the digit labels assigned to wells within another formation. We denote labelling corresponded to classes as CLASS and markup corresponded to layers in class as CLASS+LAYERS. 

\begin{table}[!ht]
\centering
\caption{Number of wells within each class of identified geological profile for considered formations}
\begin{tabular}{lcccccccccc}
\hline 
Formation           & N geological profile class         & N wells \\ 
\hline
Urenui	            &10           & 43  \\
Manganui	        &18           & 56  \\
Matemateaonga	    &8            & 45  \\
Mohakatino	        &9            & 41  \\
Moki	            &8            & 27  \\
Utsira              &5           & 13  \\
\hline 
\end{tabular}
\label{tab:labeling}
\end{table}

\begin{figure*}[!ht]
    \centering
    \includegraphics[width=0.8\linewidth]{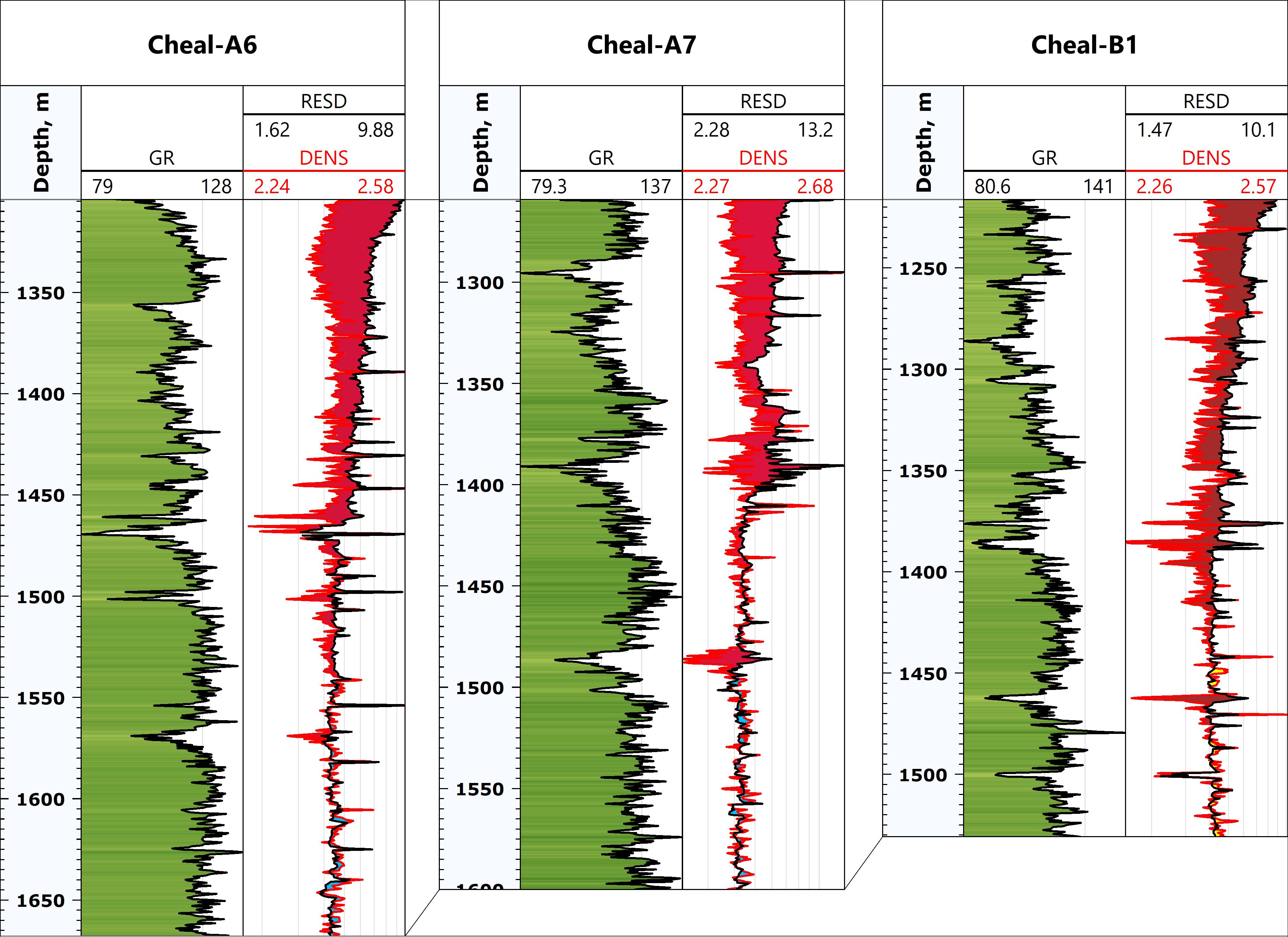}
    \caption{Example of three wells referred to the same class based on the analysis of well-logging data for the Urenui formation. Blue lines represent layer tops and bottoms determined during interwell correlation.}
    \label{fig:labelling_ex}
\end{figure*}

Obtained profile class labels give us a source for validating similarities obtained via our data-based models. An example of a pair of wells of a similar class (class 1) with dividing into layers for the Urenui formation is presented in Figure~\ref{fig:labelling_ex}.

\begin{figure}[!ht]
    \centering
    \includegraphics[width=\linewidth]{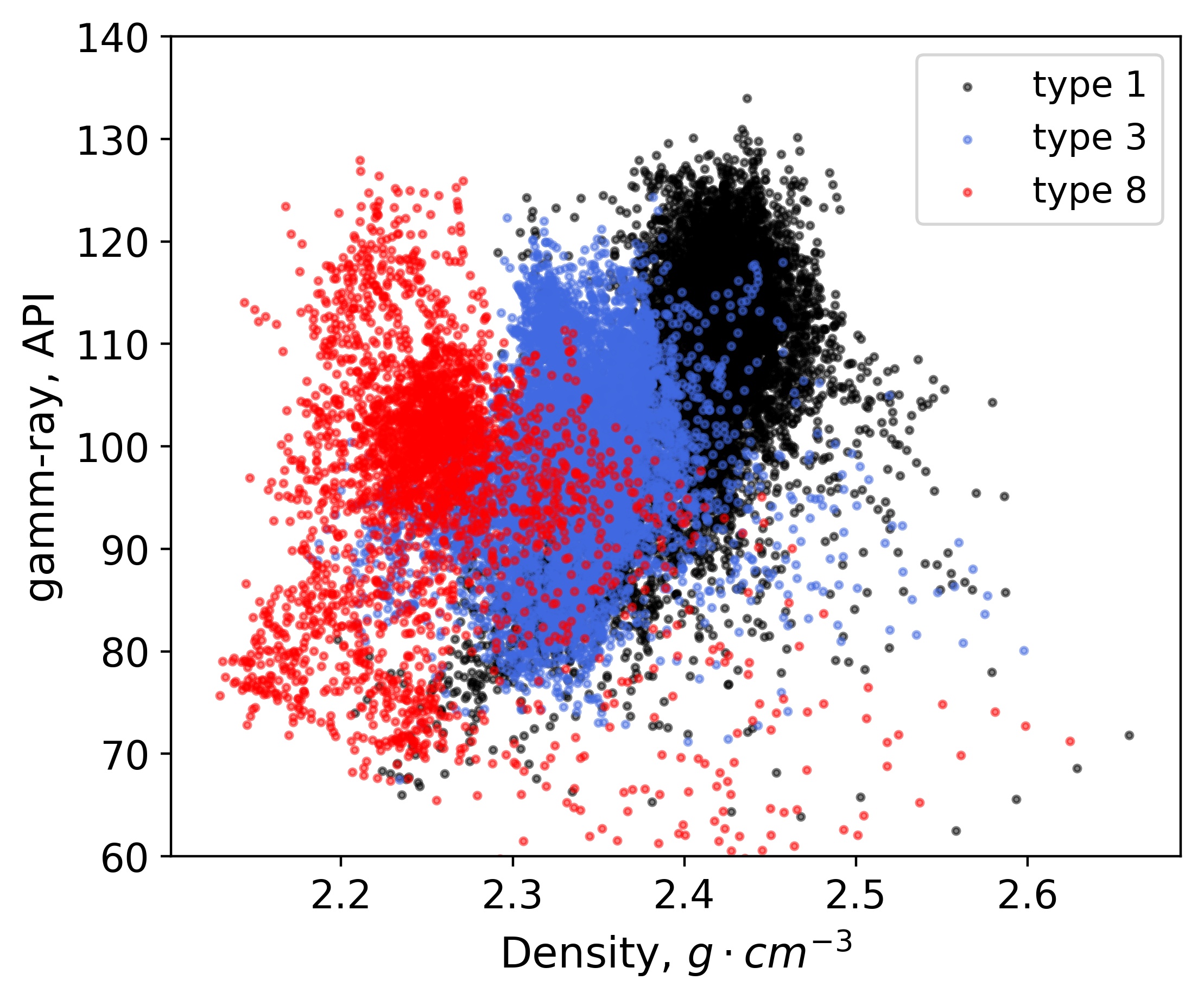}
    \caption{Example of interrelation between gamma-ray data and density log data within the Urenui formation for different types of geological profile. Each point represents a single measurement. Black colour represents the geological profile of the first class (type), blue colour --- the third class, and red colour --- the eighth class.}
    \label{fig:interrelation}
\end{figure}

Figure~\ref{fig:interrelation} plots an example of interrelation between gamma-ray data and density log data within the Urenui formation for different classes of geological profiles. As seen from Figure~\ref{fig:interrelation}, the difference in sedimentation environment even within the same formation results in distinct manners of interrelation between rock physical properties.

\subsection{Data preprocessing}\label{sec:data_preprocessing}

The complexity of oil\&gas data requires careful preprocessing. The steps in our case include (1) filling missing values, (2) dealing with specific logging measurement peculiarities and (3) selection of intervals from well required for the training of our model.
We require an additional step for baseline models that need meaningful features as inputs: (4) a feature vector generation.
Below we describe our approach to steps (1), (2), and (4).
The sampling step is crucial for the success of similarity learning,  as~\cite{hermans2017defense} suggests and requires more details, so we dedicated a separate subsection to it below. 

\emph{(1) Filling missing values.} A significant share of measurements in analysed data are missing: every feature has up to 67\% of missing values. A filling strategy may allow using most of the available data. We cope with missing values in the following way: if possible, we fill in each missing value with the previous non-missing one (forward fill), otherwise with the closest available future value (backward fill). 
We expect that as we use many intervals to evaluate the similarity between wells, these strategies will provide enough coverage for our methods to work.

\emph{(2) Dealing with peculiarities.} Machine learning and deep learning models typically require data processing due to the specificity of technological conditions in wells during logging. In our cases, preprocessing includes the following steps:
\begin{enumerate} 
 \item Eliminate physically inadequate data, i.e. intervals where electrical resistivity is less or equal to 0.
 \item Eliminate data from cavernous intervals based on the calliper data, i.e. drop rows where delta between calliper and bit size greater than 0.35 (the expert chose delta).
 \item Convert all electrical resistivity data to log-normal scale.
 \item  Fill missing values via forward and backward fill.
 \item Normalize gamma-ray and neutron log data within each well and formation by subtracting the mean and dividing by unit variance.
 \item Normalize other features by subtracting the mean and dividing by unit variance.
\end{enumerate}

\emph{(4) Feature vector generation.} As we work on interval-level, we should adapt our data for classic machine learning methods (see ~\ref{sec:methods}). A typical solution for the problem is aggregation features on intervals using, e.g. mean and std \cite{christ2018juliusneuffer, gurina2020application, romanenkova2019real}. 
So, each interval of size $(100, k)$ corresponds to the vector of features of size $2k$, where $k$ is the number of utilized features. 
The first $k$ components correspond to mean values of features.

\subsection{Sampling}

We sample pairs of intervals of a fixed length of $100$ measurements from the wells for our main experiments. 
Depth difference between two consecutive measurements is one foot.
Such techniques allow us to generate large training and validation sets of pairs of intervals that can be crucial for deep learning methods. 
For training, we generated stratified samples that have almost equal pairs of similar and different intervals.
For other parts of experiments connected with constructing a whole well representation, we cut the well to non-overlapping intervals with an equal length of $l = 100$. 
In a common classification of sampling strategies, we can call it a stratified variant of batch all sampling for similarity learning described e.g. in~\cite{hermans2017defense}.

As we use deep learning models to process the data, the interval of slightly higher or slightly shorter length can also be used as inputs to our similarity model.
The particular length $100$ was selected, as it provides reasonable quality for constructed similarity models, allows fast processing, and is big enough to capture properties based on logging data. 

For different wells we have different lengths, so sampling will give us different resolutions for each well, and we sample the same number of intervals for each well.
In our experiments, we used $100$ intervals from each well as a trade-off between well coverage and similarity calculation speed.

\section{Methods}\label{sec:methods}

We structure the methods section in the following way:
\begin{itemize}
    \item In subsection~\ref{ref:statements} we start with more formal problem statements of training models that produce data-based similarities between wells and intervals. We note that interval-level similarity allows learning without labelled data, which is crucial when resources are limited or human judgement is subjective.
    \item In subsection~\ref{subsec:clas_contrastive} we continue with the description of loss functions that we minimize during model training with respect to the model parameters. We emphasize the importance of using a contrastive learning approach, which can be run full-powered only in a representation learning framework.
    \item In subsection~\ref{subsec:ml_models} we detail on machine learning models we train with loss functions defined in the previous subsection.
    \item Then, in subsection~\ref{subsec:aggregation} we go back from the interval level similarity to the well level similarity by introducing aggregation of interval-level similarities. We also describe an alternative end-to-end approach that trains the well similarity model as a whole.
    \item We conclude with subsection~\ref{sec:domain_adaptation} that describes how we test the applicability of our models to oilfields, which is different from that used to construct an initial model.
\end{itemize}

\subsection{Problem statements}\label{ref:statements}

\paragraph{Well similarity learning.} 

Whole well comprises many different sub-layers, so with computer similarity between sub-intervals of two wells and then aggregate it, we can learn more precise patterns. Thus, instead of direct learning whole well representation, we aggregate similarity scores at the interval-level to obtain similarity scores at the well-level.  
This approach also allows the natural calculation of similarities between wells of different lengths.

After getting such similarity scores, we evaluate them according to expert labels for wells or scores.
As at the interval-level during data-based model training we don't use labelled data, our approach is unsupervised or, more precisely, semi-supervised.

\begin{figure}[h!]
    \centering
    \includegraphics[, angle =-90, width=0.975\linewidth]{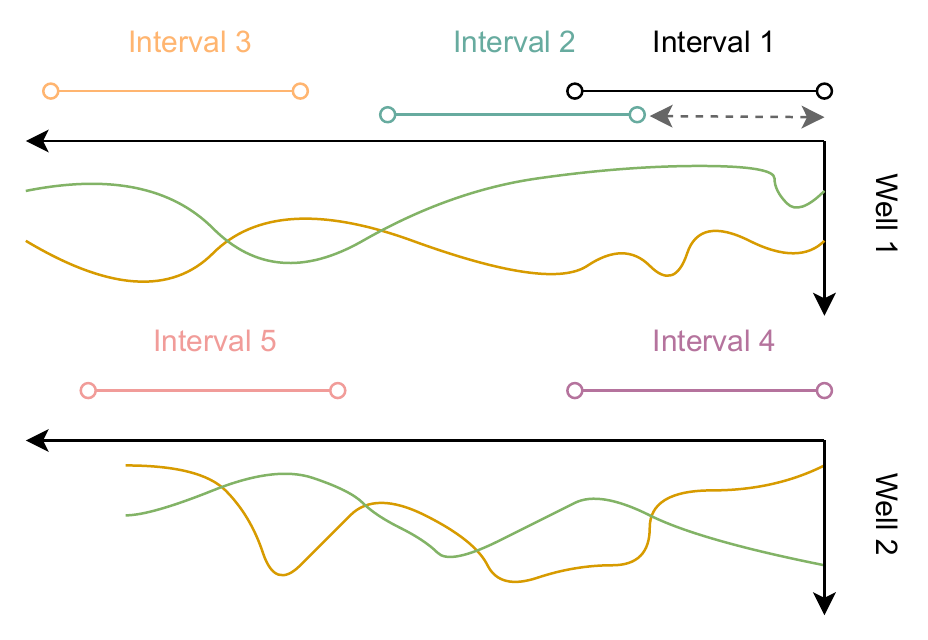}
    \caption{Well linking and Close well linking labelling example. 
    For Well linking we consider intervals in pairs (Int. 1, Int 2.) and (Int. 1, Int. 3) similar as they belong to the same well, and pairs (Int. 1, Int 4.) and (Int. 1, Int. 5) different, as intervals in these pairs belong to different wells.
    For Close Well linking we consider intervals in pair (Int. 1, Int 2.) as they belong to the same well and have small depth difference, while for (Int. 1, Int. 3) we label intervals different, as they have a large difference in depth. We still consider pairs (Int. 1, Int 4.) and (Int. 1, Int. 5) different, as intervals in these pairs belong to different wells.}
    \label{fig:close_well_linking}
\end{figure}

\paragraph{Well interval similarity learning.} 

We consider a dataset that consists of well's interval $X_i = (x_{1i}, \ldots, x_{l_i i}), x_{ji} \in \mathbb{R}^{d}$, where $l_i$ is the $i$-th interval length, and $d$ is the number of features collected at each step of an interval. 
Our goal is to construct a data-based model that takes as input data about two intervals $(X_i, X_j)$ and outputs a similarity between two intervals $s_{ij} = s(X_i, X_j)$.
If intervals are close to each other, the similarity model should output values close to $1$, otherwise, evaluate similarity should be close to $0$.

We consider two approaches to label the interval pairs to train a similarity classifier:
\begin{itemize}
    \item For \emph{Linking problem}, we expect the classifier to output $1$ if two intervals come from the same well and $0$ if they come from two different wells.
    \item For \emph{Close linking problem}, we expect the classifier to output $1$ if two intervals come from the same well \emph{and are close to each other in terms of depth} (starts of the intervals differ no more than $50$ measurements) and $0$ if they don't meet these two requirements.
\end{itemize}

An example of our labelling for different pairs is given in Figure~\ref{fig:close_well_linking}.

\begin{figure*}[!ht]
    \centering
    \includegraphics[width=0.8\linewidth]{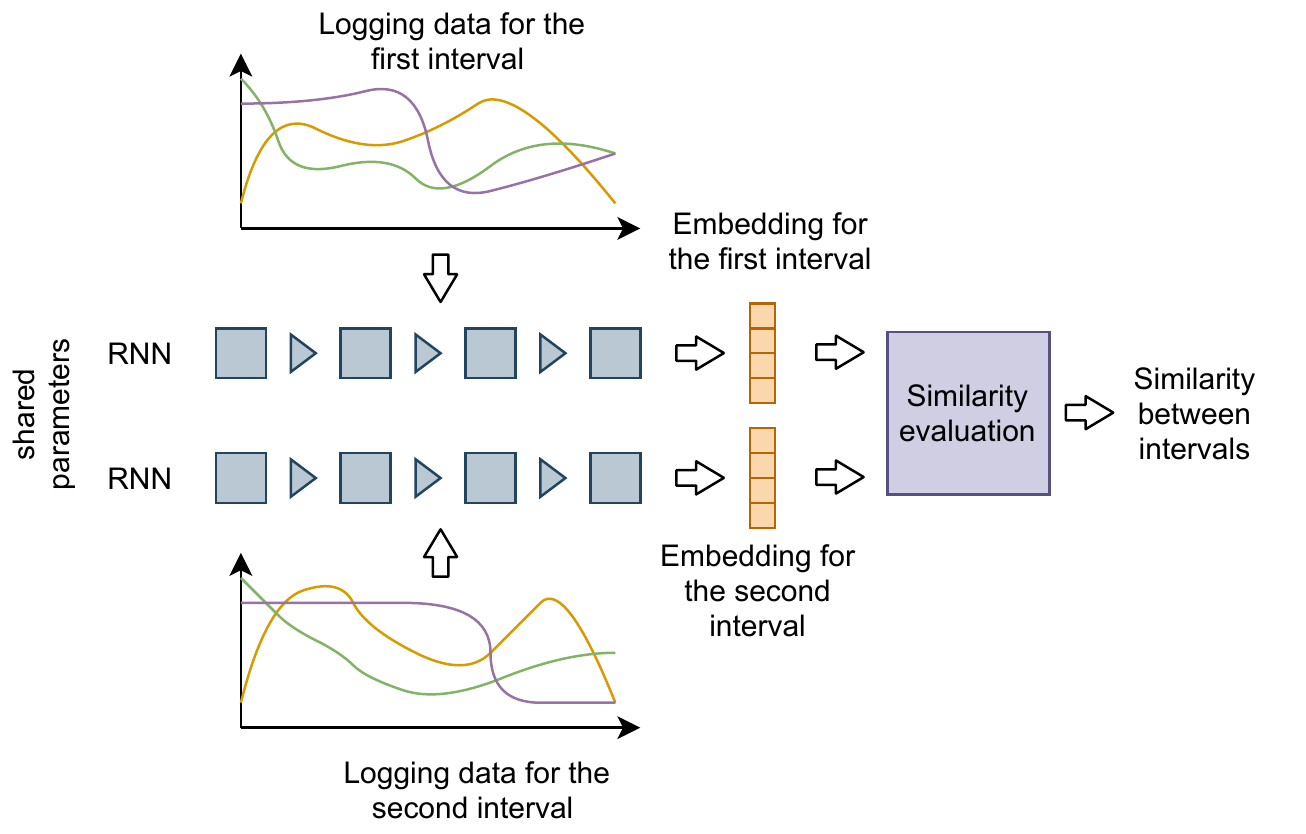}
    \caption{Similarity evaluation model for a pair of intervals. We run Recurrent Neural Network (RNN) two times to process data from two intervals and obtain their embeddings. To evaluate similarity from a pair of embeddings, we then use a simple formula or a simple layers neural network. During training, we modify parameters of the RNN and Similarity evaluation part to match the target similarities between pairs of intervals.}
    \label{fig:siam_nn}
\end{figure*}

\begin{figure*}[!ht]
    \centering
    \includegraphics[width=0.6\linewidth]{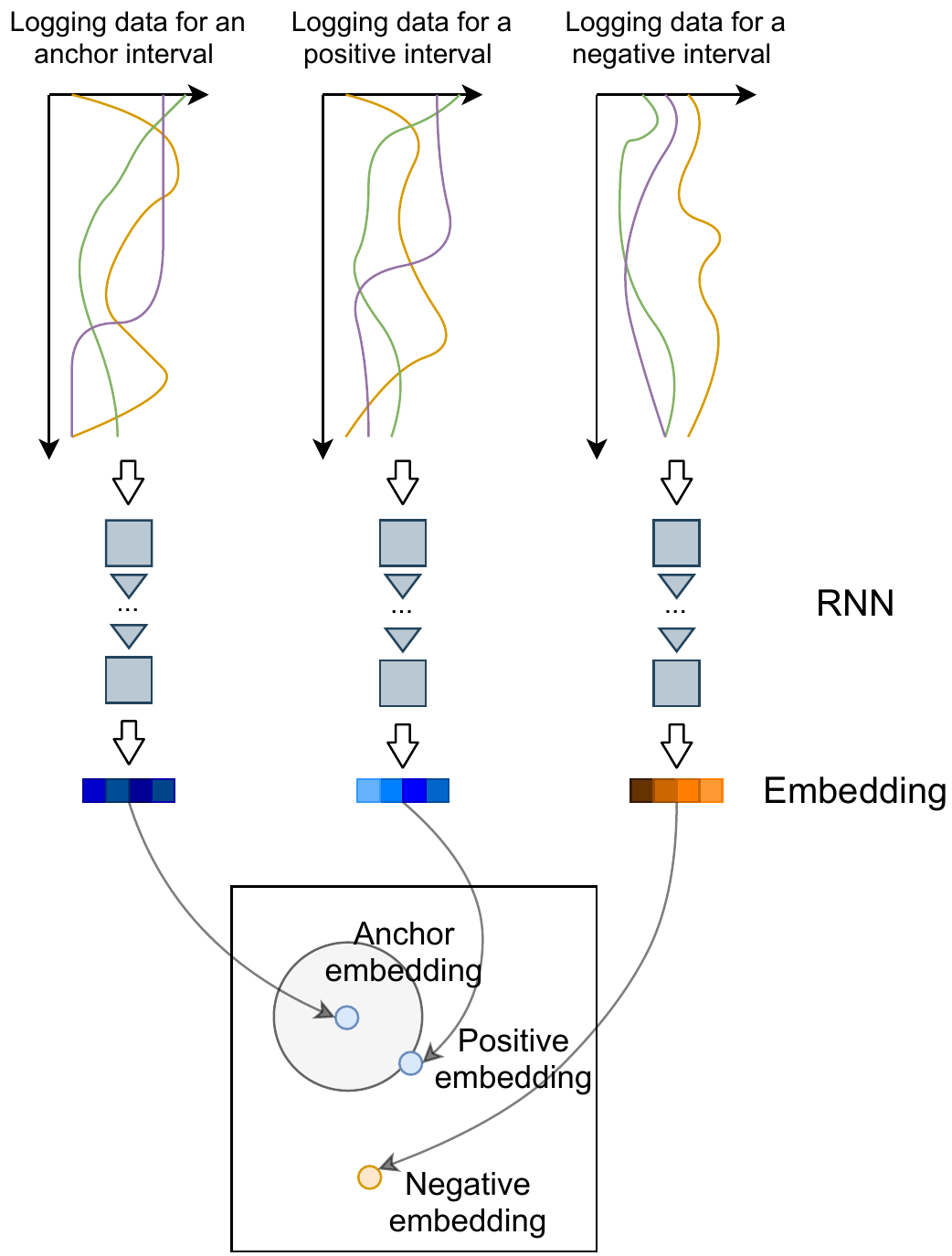}
    \caption{Triplet loss intuition. We consider a triplet of interval objects: an anchor, a positive object similar to the anchor, and a negative object dissimilar from the anchor. After embedding triplet intervals, we penalize the model if the distance between the anchor and the negative is smaller than the distance between the anchor and the positive.}
    \label{fig:triplet_model}
\end{figure*}

\subsection{Classification-based and contrastive-based approaches to similarity learning} \label{subsec:clas_contrastive}

This work is devoted to developing a representation-based similarity model in this work. Such an approach allows operating with different data types and transferring obtained embedding to other tasks, e.g., detecting anomalous patterns~\cite{jing2020self}.

To obtain similarities, we follow a common paradigm for state-of-the-art deep learning based on a two-phases model. Figure~\ref{fig:siam_nn} represents a general scheme of how we obtain similarities between two intervals evaluated by our model. We apply the same \emph{encoder} model $E_i = f(X_i)$ to get embeddings for each interval. Then we compare embeddings using an additional procedure $s_{ij} = g(E_i, E_j)$ with no or a small number of parameters to estimate during training. The output of this part reports similarity between wells' intervals.

There are two approaches to train such models: one method is to force the model to solve \emph{classification problem} (similar or not objects); another is to move representations directly in the latent space. We use a fully-connected model or geometrical distance to estimate similarities for the first approach. This way allows achieving better metrics from machine learning points of view. However, no guarantee that obtained representations are located in a latent space as we expected, i.e. embeddings that corresponded to similar objects lie closer to each other and different embeddings are separated. 

A natural alternative to operate with representations directly is to use \emph{contrastive learning} depicted at~\ref{fig:triplet_model}. We consider a set of triplets; each triplet consists of an anchor interval $X_a$, a positive interval $X_p$ and a negative interval $X_n$. We select them such that the anchor interval and the positive interval should be similar according to the similarity measure. Simultaneously, the anchor and negative intervals should be far from each other. Formally, the triplet loss equals $\max(d(X_a, X_p) - d(X_a, X_n) + \alpha, 0)$, where $\alpha$ is a margin: if $d(X_a, X_n)$ is significantly (by more than $\alpha$) smaller, than $d(X_a, X_p)$, then we penalize the calculated similarities and force our model to avoid such situations. We average such loss over a set of triplets.

In contrast to the classification-based approach above, \emph{triplet model}  based on triplet loss doesn't solve Linking problems directly. Thus, we need a classifier over the triplet model. To calculate similarities, distance, we evaluate Euclidean \eqref{ed_dist} or cosine distance \eqref{cos_dist} between obtained embeddings from triplet model.

To sample such triplets, we need some definite criteria for choosing similar/dissimilar objects~\cite{babaev2020event}. We sample triplets so that for each anchor and positive belonging to the same well, there are several negatives from each different well.

\subsection{Machine learning models}\label{subsec:ml_models}

\paragraph{Classic Machine Learning methods.} 
As three non-rep\-resentation learning baselines, we use the naive geometric-based distance between features, Logistic regression and random forest methods.

A simple baseline is to estimate similarity based on Euclidean $s(X_i, X_j)_{ED}$ or cosine $s(X_i, X_j)_{cos}$ distance between intervals. 
In this case, the formulas for similarity are following:
\begin{equation}\label{ed_dist}
    s(X_i, X_j)_{ED} = \frac{1}{1 + ||X_i - X_j||_2},
\end{equation}

\begin{equation}\label{cos_dist}
    s(X_i, X_j)_{cos} = \frac{(X_i \cdot X_j)}{||X_i||_2 ||X_j||_2}
\end{equation}

Logistic Regression is a generalized linear model for the classification problem~\cite{friedman2001elements}. The central assumption is that the probability for an object to belong to a target class is a sigmoid monotonic transformation of a linear weighted sum of input features. In our experiments, we use Logistic Regression with default hyperparameters. 
This method is robust and fast to train if the features' number is relatively small compared to the data size.
On the other hand, linearity assumption is too strong for most problems, so typically, better approaches for tabular data exist. 

The random forest of decision trees model is the most popular solution for regression and classification machine learning problems~\cite{friedman2001elements}. Its combination with Gradient Boosting, a specific method for the training, allows overcoming many technical difficulties. For example, such models can easily handle data of various sample sizes and quality, automatically process missing data, learn quickly with a large number of features~\cite{kozlovskaia2017deep, romanenkova2019real} and are suitable in different settings~\cite{ke2017lightgbm,dorogush2018catboost}. We use a gradient boosting realization XGBoost with default hyperparameters~\cite{chen2015xgboost}. 

Although we can use classic machine learning approaches, they don't produce any representations. As we see a lot of new types of geological data~\cite{deng2019arcface}, it is more convenient to obtain similarity during representation learning and apply neural-based models for it. 

\paragraph{Neural Networks.}

Neural networks are the current state-of-the-art for the representations learning~\cite{goodfellow2016deep}. The most natural choice for sequential data is model architectures based on recurrent neural networks (RNNs) ~\cite{jozefowicz2015empirical}. For the encoder model, i.e. to obtain representations,  we use LSTM (Long short-term memory), one of the best types of RNNs~\cite{greff2016lstm}. LSTM can capture temporal dependencies and handle long-range dependencies that appeared in the data~\cite{gers2000learning, greff2016lstm}. The model proceeds well-logging data along its depth sequentially one by one observation at a specific depth. There are more recent approaches like transformers and one-dimensional convolution neural networks. However, for the available amount of data and its length, they typically produce results similar to LSTM or show inferior performance~\cite{babaev2020event}.

We analyse two options to evaluate similarity. The first option is to calculate Euclidean~\eqref{ed_dist} or cosine~\eqref{cos_dist} distance 
as in baseline, but now to do it between embeddings to get similarity. The second option is to train a separate fully connected (FC) neural network to estimate similarity. The advantage of the first approach is the direct evaluation of distance, while in many cases, additional fully-connected projection layers allow better representations at intermediate neural network layers~\cite{chen2020simple}.

As we've mentioned before, we consider two approaches for training neural networks: classification-based and cont\-ras\-ti\-ve-based approaches. Accordingly, we utilize Siamese loss and Triplet loss functions. For Siamese loss, we directly minimize cross-entropy or similarity classification quality analogous to the classic machine learning method training. For Triplet loss, we use the triplet loss defined in Subsection~\ref{subsec:clas_contrastive} such that we want an anchor interval and a positive interval to be close to each other and an anchor interval and a negative interval to be far from each other.

\subsection{Aggregation methods}\label{subsec:aggregation}

In our problem, it is essential to transfer similarity from the interval-level to the well-level via a proper aggregation strategy, as, in the end, we need similarities between wells, not well intervals. 

To use interval-level similarity, we follow the procedure below.
We split $m$-th well to $k_m$ consecutive intervals with data $X^{m}_{j}$, $j = \{1, ..., k_m\}$ and the same can be done for $n$-th well. For every such interval, we obtain embedding $E^m_{j} = f_{\mathbf{w}}(X^m_{j}) \in \mathbb{R}^d$. At the same time, for the pair of intervals $(X^{m}_{j}, X^{n}_i)$, similarity model provide a similarity score $s_{ij}$. 

We apply \textbf{\textit{macro}} aggregation if we aggregate embedding from the interval in one well's embedding and then obtain a similarity score for pair of wells. 
For example, we use mean value for each component of the embeddings vector: $E^m_{i} = \frac{1}{d} \sum_{j = 1}^{k_m} E^m_{ij}$.
Then, we evaluate similarity by our model $s_{mn} = g(E^m, E^n)$.
The other option is \textbf{\textit{micro}} aggregation, i.e. we firstly calculate similarities for a pair of intervals $s_{ij} = s(X^m_i, X^n_j)$ and then aggregate it to wells' similarity $s_{mn} = \frac{1}{k_m k_n} \sum_{ij} s_{ij}$. 
To speed calculations, we operate with only part of possible pairs of intervals to get the matrix of similarities without using the excessive sum.
We randomly sample $\min($1000$, \text{available number of})$ intervals for each well in the dataset.

\subsubsection{End-to-end training}

An end-to-end training is a general name of the approaches in which one model learns to solve a complex task directly instead of dividing the task into several parts with independent training. Regarding our research topic, we investigate a model that receives logging data for wells as input and outputs the whole wells' embedding. 
This embedding can be used to calculate similarities.

We call the part corresponding to the embedding of the initial logging data as \textit{the encoder model} and the part corresponding to aggregation of intervals' embeddings as \textit{the aggregation model}. We train the aggregation model to predict the expert's labelling of wells. 

More formally, a one step for the final end-to-end training procedure is following:
\begin{enumerate}
\item make one optimization step for the intervals' representation model (any of the above). 
\item cut training wells to intervals with suitable length;
\item fed these cutting intervals into obtained representation model to get a set of embeddings;
\item input a sequence of embedding to an aggregation model and train this model to predict well's CLASS;
\end{enumerate}
We repeat these steps until convergence, optimizing the encoder and the aggregation model jointly. 

\subsection{Domain adaptation}\label{sec:domain_adaptation}

Our similarity model should extrapolate well to be useful.
We consider two properties related to the extrapolation ability of the model: (1) applicability in a wide range of scenarios and (2) a quick adaptation to data from new domains via transfer learning or without it. During our research, we analysed the data from two distinct datasets that represent different sedimentation environments, lithological composition, petrophysical properties, etc.
The first dataset comprises data from New Zealand, and the second one is on data from Norway.
In this way, we'll be able to understand the abilities of our models to adapt to a new domain.

We compare the adaptation capabilities of our model in several common ways~\cite{wang2018deep}:
\begin{itemize}
    \item Keep the model unchanged and apply it directly to new data, as we expect that new data resemble the old one. It is the fastest way, which in practice can work poorly due to differences in data collection and organization as well as due to different properties of different oilfields.
    \item Train a new model from scratch using only new data. Unlike the previous one, it is the most undesirable approach, which requires a lot of time and computational resources. Moreover, if the new data size is limited, we end up with an under-performing model.
    \item Fine-tune a bit an old model from scratch using new data to capture peculiarities of new data, which is a good compromise solution.
\end{itemize}    

If an unchanged model or a fine-tuned model outperforms a model trained from scratch using new data, we can prove that our embeddings (representations) and similarity measure are universal and can be applied in different scenarios for different oilfields.

\section{Evaluation methodology} \label{sec:metrics}
% Quality metrics

Our machine learning problem falls to either classification or regression problem statements.
For the classification problem, a data-based model predicts a class label for an object and make as few errors as possible, e.g. if a pair of intervals (object) is similar or not (two possible class labels). 
For the regression problem, a data-based model predicts a continuous variable, e.g. a distance between a pair of intervals.

Below we provide definitions and intuition behind quality metrics for each of these machine learning problem statements. The summary of metrics is in Table~\ref{tab:metrics_summary}.

\begin{table}[h!]
    \centering
    \begin{tabular}{ccc}
    \hline
    Metric & Problem & Range \\
    name     & statement & (Worst, Best)\\
    \hline
    Accuracy & Classification & (0, 1) \\
    ROC AUC & Classification & (0, 1) \\
    PR AUC & Classification & (0, 1)\\
    % $R^2$ & Regression & (0, 1) \\
    ARI & Clustering & (-1, 1) \\
    AMI & Clustering & (-1, 1) \\
    V-measure & Clustering & (0, 1) \\
    
    \hline
    \end{tabular}
    \caption{Machine learning quality metrics. We want to \underline{maximize} all metrics.}
    \label{tab:metrics_summary}
\end{table}

\subsection{Classification and regression quality metrics}
\label{sec:classification_metrics}

To evaluate the quality of the models suitable for the classification tasks, e.g. Linking from \ref{sec:methods}, we use the following classic machine learning metrics: accuracy, area under the Receiver Operating Characteristic (ROC) curve or ROC AUC and area under the Precision-Recall (PR) curve or PR AUC. 

We consider a dataset $D = {(X_i,y_i)}_{i = 1}^N$. In this notation, $X_i \in \mathbb{R}^{l \times d}$ is an interval with length $l$ consisting of $d$-dimensional measurement at every step and $y_i$ is a true value of the target. For the Linking problem described in the \ref{sec:methods}, $y_i$ is either 0 (dissimilar intervals) or 1 (similar intervals). Denote $\hat{y_i} = f(X_i)$ a predicted label by the classifier model $f(X)$. 

The most natural quality metric is accuracy.
The accuracy is the share of correct predictions: 
\[
Accuracy = \frac{1}{N} \sum_{i = 1}^N [y_i = \hat{y_i}],
\]
where $[e]$ is the indicator function, which equals one, if the condition inside the brackets holds and equals zero otherwise.

To define ROC AUC and PR AUC, we firstly introduce more basic metrics reflected in the so-called confusion matrix that represents a classification model performance. The confusion matrix consists of four metrics: [Number of] True Positive (TP), False Negative (FN), False Positive (FP), True Negative (TN) objects:

\begin{align*}
    TP &= 1/N \sum_{i = 1}^N [y_i = 1][\hat{y_i} = 1], \\ 
    FN &= 1/N \sum_{i = 1}^N [y_i = 1][\hat{y_i} = 0], \\
    FP &= 1/N \sum_{i = 1}^N [y_i = 0][\hat{y_i} = 1], \\
    TN &= 1/N \sum_{i = 1}^N [y_i = 0][\hat{y_i} = 0].
\end{align*}
These metrics distinct correct labels and errors for the first and for the second class in our classification problem. 

We can also regard some related metrics true positive rate (TPR) or Recall, false positive rate (FPR) and Precision:
\[TPR = Recall = \frac{TP}{TP + FN}\]
\[FPR = \frac{FP}{FP + TN}\]
\[Precision = \frac{TP}{TP + FP}\]

It's important to note that actually, every classification model predicts not a particular label but the probabilities $p$ of belonging to the class $p(X) = f_p(X) \in [0, 1]$. To obtain a label, the model compares received probabilities with some threshold $p_0$. We suppose that an object belongs to the class if the corresponded predicted probability is above a threshold $p > p_0$. 

By comparing obtained probabilities with a set of thresholds from 0 to 1 and calculating confusion matrices for each threshold, we get a trajectory in the space (TPR, FPR) and similar in the space (Precision, Recall) with each pair of values corresponding to a specific threshold. Received curves are named ROC curve and Precision-Recall curve correspondingly, and the used metrics are areas under these curves. 

\subsection{Clustering quality metrics}\label{sec:cluster_metrics}

Clustering quality metrics serves for the comparison of two labelling of objects if the exact ordering is either absent or not important.

In our case, the first labelling is the true labelling corresponding to expert marks for wells or well intervals. The second labelling corresponds to cluster labels obtained after clustering of the embeddings of wells or intervals provided by our data-based encoder model in our similarity model. 

If cluster labels are similar to true labels, the evaluated representations and model generated them are good. Otherwise, we report our embeddings to be poor, given true expert labels.
We consider three types of target values: 
\begin{itemize}
    \item WELLNAME --- we expect that embedding of intervals from one well will be close to each other. This labelling doesn't use any expert judgement; 
    \item CLASS --- we expect that embedding of intervals from one class (expert's labelling) will be close to each other, but embeddings can match to different wells and lie in different layers (expert's labelling);
    \item CLASS + LAYER --- we expect that embedding of intervals from one class and one layer (expert's labelling) will be close to each other, but embeddings can match to different wells; 
\end{itemize}

\begin{figure}
    \centering
    \includegraphics[width=\linewidth]{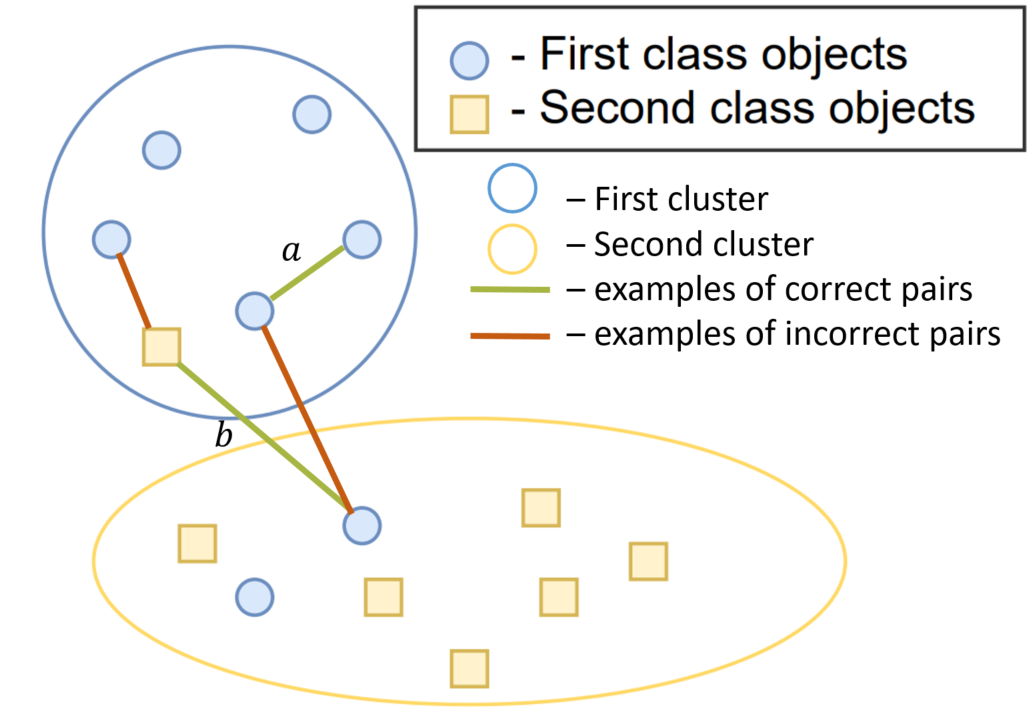}
    \caption{The intuition behind the comparison of labels provided as class labels and labels via clustering. In this example, we have two possible classes for objects and two possible clusters. Colours of circles correspond to true classes of objects, and ellipsoids are proposed clustering of objects. We want them to match and compare how the distribution of pairs between proposed clusters and true class labels correspond. We want the number of correct pairs to be high and the number of incorrect pairs to be low.}
    \label{fig:rand_index}
\end{figure}

The intuition behind our validation methodology is in Figure~\ref{fig:rand_index}. 

We use Adjusted Rand Index (ARI) as the clustering quality metric~\cite{rand1971objective}, as it provides good insight on clustering quality and has an intuitive explanation behind it. 
Adjusted Rand Index (ARI) measures the similarity of the two assignments, ignoring permutations of the order of possible labels. 

A formal definition of ARI is the following. Let's start with a simple Rand index without adjustment. Denote $n$ the total number of objects from the validation sample, $a$ the number of pairs in the same cluster and have the same label, $b$ the number of pairs that are in different clusters and have different labels.
Then Rand Index (RI) is:
\[
    {\rm RI} = \frac{a + b}{C_{2}^{n}},
\]
where $C_2^{n} = \frac{n(n-1)}{2}$ is the binomial coefficient and equals to the total number of distinct pairs for a sample of size $n$.
At the same time, the range of possible values for the Rand index is huge, and a particular value doesn't designate a good or bad result.

The Adjusted Rand Index is Rand Index corrected for the chance of random label assignments:
\[
    {\rm ARI} = \frac{{\rm RI} - \mathbb{E}[{\rm RI}]}{\max ({\rm RI}) - \mathbb{E}[{\rm RI}]},
\]
where $\mathbb{E}[{\rm RI}]$ is the expectation of Rand index for random assignment of cluster labels.
ARI takes values from $-1$ to $1$.
Value $0$ corresponds to a random label assignment, value $1$ corresponds to the perfect assignment, and negative values describe cases worse than random assignment.

As an alternative, we consider Adjusted Mutual Information (AMI) quality metric.

Assume two label assignments $U$ and $V$. Denote $P_i = |U_j| / n$ and $P'_j = |V_j| / n$. Then the entropy of $U$ is: 
\[ H(U) = - \sum_{i=1}^{|U|} P_i \log P_i \] 
Similarly for $V$. Denote $P_{ij} = |U_i \cap V_j| / n$. Then the Mutual Information (MI) between $U$ and $V$ is: 
\[
{\rm MI}(U, V) = P_{ij} \log \left( \frac{P_{ij}}{P_i P'_j} \right)
\]

The Adjusted Mutual Information is Mutual Information adjusted for the chance of random label assignments:
\[
{\rm AMI} = \frac{{\rm MI} - \mathbb{E}[{\rm MI}]}{\operatorname{mean} (H(U), H(V)) - \mathbb{E}[{\rm MI}]},
\]
again $\mathbb{E}[{\rm MI}]$ is the expected mutual information for a random assignment.

Finally, we consider the V-measure that is the harmonic mean of homogeneity $h$ and completeness $c$. Homogeneity is the property of cluster assignment that each cluster contains only members of a single class:
\[
h = 1 - \frac{H(C|K)}{H(C)}
\]
Completeness is the property of cluster assignment that all members of a given class are assigned to the same cluster:
\[
h = 1 - \frac{H(K|C)}{H(K)}
\]
Here, $H(C|K)$ is the conditional entropy of the classes given the cluster assignments and $H(C)$ is the entropy of the classes:
\[
H(C|K) = - \sum_{c=1}^{|C|} \sum_{k=1}^{|K|} \frac{n_{ck}}{n} \log \left( \frac{n_{ck}}{n} \right),
\]
$n_{ck}$ is the number of objects from $c$-th cluster that fall into $k$-th class.
\[
H(C) = - \sum_{c=1}^{|C|}\frac{n_{c}}{n} \log \left( \frac{n_{c}}{n} \right)
\]
The V-measure is then:
\[
v = 2 \cdot \frac{h \cdot c}{h+c}
\]

In all our experiments, these metrics are obtained by Agglomerative clusterisation~\cite{murtagh2014ward} with default parameters as, in our opinion, it is the most sustainable algorithm suitable for our task and experiments suggested that it works better in our scenario compared to other clustering algorithms. In contrast to other clustering methods, results for Agglomerative clustering are similar for different runs because there is no randomization in this algorithm.

\section{Results and discussions}\label{sec:results}

Our results section has the following structure:

\begin{itemize}
    \item We provide the quality of similarity models in terms of ML metrics and expert's labelling in Subsection~\ref{sec:sim_models_results}.
    
    \item Additional sanity checks for our models are given in Subsection~\ref{sec:sanity_checks}.
    \item End-to-end deep learning model training results are in Subsection~\ref{sec:end2end_results}. We compare them to our self-supervised approach that doesn't need expert labelling
    \item We finalize our results with domain adaptation quality evaluation in Subsection~\ref{sec:transfer_learning_results}.
\end{itemize}

If not specified otherwise, we use wells from Urenui formation for training and testing of data-based models. 

\subsection{Similarity models quality}\label{sec:sim_models_results}

In this subsection, we present the quality of similarity models that identify if two intervals from wells are similar or not for two problem statements \emph{Linking problem} and \emph{Close linking problem} (see ~\ref{sec:methods} for more details).
As basically it is a classification of pairs problem, we report classification quality metrics.

\subsubsection{Experiment setup}

We use cross-validation to access the model quality: we split the whole dataset fixed number of times to train and test samples, train sample is used for model training, a test sample is used for the quality metrics evaluation. We presented mean values of obtained metrics for all splits as well as standard deviations for them.

For our validation scheme, all data from a well fall completely into either the test or training sets. Then we train the model using all other wells and get predictions for hold-out wells.

Possible model design choices include variability of the model architecture. We carefully experiment with them to identify the best combination that will serve for more advanced comparisons.

The different data-based models used in these experiments are the following:
\begin{itemize}
    \item Classic methods
    \begin{itemize}
        \item \textit{Logistic regression}: default parameters;
        \item Gradient boosting (\textit{XGBoost}): default parameters; 
        \item \textit{Euclidean distance};
        \item \textit{Cosine distance};
        
    \end{itemize}
    \item Neural Networks 
    \begin{itemize}
        \item \textit{Siamese + 3 FC.} The exact model architecture is LSTM with the hidden size 16 + FC (32, 64) + ReLU + Dropout (0.25) + FC (64, 32) + ReLU + Dropout (0.25) + FC (32, 1) + Sigmoid;
        \item \textit{Siamese + X distance}, where X denotes Euclidean, Cosine. The exact model architecture is LSTM with hidden size 16 + distance calculation;
        \item \textit{Triplet + X distance}, where X denotes Euclidean, Cosine. The exact model architecture is LSTM with hidden size 64 + distance calculation;
    \end{itemize}
\end{itemize}

For training neural networks, we use Adam (see ~\cite{kingma2014adam}) optimizer with a learning rate of $0.001$.

\subsubsection{Results and discussion}

The comparison of the best models for each technique in terms of machine learning metrics is in Table~\ref{tab:close_well_linking} and ~\ref{tab:well_linking}. The results are averaged over $5$ five (you normally write numbers till 10 with words) folds. Logistic regression performs worst of all, so simple models are not good enough for the problem at hand. XGBoost variant of Gradient boosting provides better qualities, while Recurrent Neural Networks shows superior performance in case of \emph{Close linking} and comparable performance in case of \emph{Linking} problem. 

\begin{table*}[!ht]
\centering
\footnotesize
    \begin{tabular}{lccccc}
    \hline
	Metrics &Accuracy &Precision &Recall &ROC AUC &PR AUC \\\hline
	Logistic Regression &0.725 $\pm$ 0.074 &0.694 $\pm$ 0.073 &0.608 $\pm$ 0.200 &0.509 $\pm$ 0.092 &0.490 $\pm$ 0.082 \\
	XGBoost &\textbf{0.977 $\pm$ 0.014} &\textbf{0.970 $\pm$ 0.025} &0.775 $\pm$ 0.053 &0.824 $\pm$ 0.041 &0.954 $\pm$ 0.028 \\
	Euclidean distance &0.552 $\pm$ 0.111 &0.652 $\pm$ 0.135 &0.764 $\pm$ 0.037 &0.295 $\pm$ 0.049 &0.601 $\pm$ 0.148 \\
	Cosine distance &0.663 $\pm$ 0.046 &0.831 $\pm$ 0.048 &0.651 $\pm$ 0.038 &0.728 $\pm$ 0.053 &0.866 $\pm$ 0.043 \\
	Siamese + Eucl. dist. &0.700 $\pm$ 0.114 &0.700 $\pm$ 0.114 &\textbf{1.000 $\pm$ 0.000} &0.814 $\pm$ 0.056 &0.894 $\pm$ 0.071 \\
	Siamese + Cos. dist. & 0.804 $\pm$ 0.076 &0.834 $\pm$ 0.110 &\ul{0.900 $\pm$ 0.062} &0.874 $\pm$ 0.064 & 0.930 $\pm$ 0.051 \\
	Siamese + 3 FC &0.842 $\pm$ 0.052 &0.955 $\pm$ 0.035 &0.793 $\pm$ 0.079 &\textbf{0.934 $\pm$ 0.047} &\textbf{0.951 $\pm$ 0.041} \\
	%Siamese + Euclidean + 3 FC &0.960 $\pm$ 0.029 &0.966 $\pm$ 0.028 &0.644 $\pm$ 0.043 &0.735 $\pm$ 0.040 &0.939 $\pm$ 0.036 \\
	%Siamese + Cosine + 3 FC &\ul{0.961 $\pm$ 0.029} &0.932 $\pm$ 0.046 &0.872 $\pm$ 0.115 &\ul{0.880 $\pm$ 0.076} &\ul{0.941 $\pm$ 0.040} \\
	Triplet + Eucl,dist. & \ul{0.914 $\pm$ 0.041} &\ul{0.968 $\pm$ 0.019} &0.431 $\pm$ 0.123 &0.708 $\pm$ 0.061 &0.909 $\pm$ 0.048 \\
	Triplet + Cos.Dist. &0.911 $\pm$ 0.042 &0.723 $\pm$ 0.060 &0.898 $\pm$ 0.083 &0.770 $\pm$ 0.042 &0.903 $\pm$ 0.046 \\
	\hline
    \end{tabular}
    \caption{Comparison of the quality of models for Linking problem. For each quality metric, TOP-1 values are highlighted with \textbf{bold} font and TOP-2 best values are \underline{underlined}.}
    \label{tab:well_linking}
\end{table*}

\begin{table*}[!ht]
\centering
\footnotesize
    \begin{tabular}{lccccc} % (?)
    \hline
	Models &Accuracy &Precision &Recall &ROC AUC &PR AUC \\ \hline
	Logistic Regression &0.483 $\pm$ 0.089 &0.523 $\pm$ 0.143 &0.213 $\pm$ 0.145 &0.682 $\pm$ 0.066 &0.647 $\pm$ 0.062 \\
	%Regression & & & & &\\
	XGBoost &0.787 $\pm$ 0.042 &\ul{0.752 $\pm$ 0.039} &0.575 $\pm$ 0.078 &\ul{0.802 $\pm$ 0.041} &\ul{0.895 $\pm$ 0.032} \\
	Euclidean distance &0.300 $\pm$ 0.015 &0.287 $\pm$ 0.027 &0.680 $\pm$ 0.015 &0.310 $\pm$ 0.034 &0.266 $\pm$ 0.020 \\
	Cosine distance &0.596 $\pm$ 0.048 &0.449 $\pm$ 0.058 &0.697 $\pm$ 0.051 &0.674 $\pm$ 0.045 &0.530 $\pm$ 0.049 \\
	Siamese + Eucl. dist. &0.348 $\pm$ 0.031 &0.348 $\pm$ 0.031 & \textbf{1.000 $\pm$ 0.000} &0.663 $\pm$ 0.039 &0.510 $\pm$ 0.043 \\
	Siamese + Cos. dist. &0.603 $\pm$ 0.061 &0.458 $\pm$ 0.065 &0.749 $\pm$ 0.084 &0.710 $\pm$ 0.070 &0.592 $\pm$ 0.076 \\
	Siamese + 3 FC &\ul{0.814 $\pm$ 0.053} &0.742 $\pm$ 0.103 &0.707 $\pm$ 0.082 &\textbf{0.874 $\pm$ 0.056} &0.766 $\pm$ 0.126 \\
%	Siamese + Euclidean + 3 FC &0.684 $\pm$ 0.112 &0.651 $\pm$ 0.062 &0.707 $\pm$ 0.093 &0.769 $\pm$ 0.038 &0.836 $\pm$ 0.048 \\
%	Siamese + Cosine + 3 FC &0.647 $\pm$ 0.110 &0.539 $\pm$ 0.028 &0.846 $\pm$ 0.099 &0.695 $\pm$ 0.014 &0.813 $\pm$ 0.045 \\
	Triplet + Euclidean &\textbf{0.925 $\pm$ 0.024} &\textbf{0.963 $\pm$ 0.027} &0.495 $\pm$ 0.142 &0.737 $\pm$ 0.068 &\textbf{0.925 $\pm$ 0.024} \\
	Triplet + Cosine &\textbf{0.926 $\pm$ 0.022} &0.714 $\pm$ 0.040 &\ul{0.952 $\pm$ 0.014} &0.783 $\pm$ 0.036 &\textbf{0.926 $\pm$ 0.020} \\
	
	\hline
    \end{tabular}
    \caption{Comparison of the quality of models for Close linking problem. TOP-1 values are highlighted with \textbf{bold} font and TOP-2 best values are \underline{underlined}.}
    \label{tab:close_well_linking}
\end{table*}

The results show that the best model judging by the ROC AUC score is the Siamese model with 3 (three ?) fully-connected layers, but considering PR AUC Triplet models with the distance perform better. So, we expect that the representations obtained with deep neural networks given insides on well intervals and overall well characteristics. 

\subsection{Representation analysis}
\label{sec:representation_results}

To compare the obtained embeddings from our best models Siamese + 3FC and Triplet, we clustered representations obtained from them by clusters based on WELLNAME, CLASS, CLASS+LAYER targets and evaluated corresponding metrics. 
We are considering clusterization for embeddings of intervals. Below we investigate similar results for the whole-well embeddings. 
As in the case of classic machine learning metrics above, the results are also averaged over $5$ five folds. 

Tables \ref{tab:well_linking_clustering} and \ref{tab:close_well_linking_clustering} demonstrate the results for considered Neural network based models outperform with the clusterisation based on original features without embedding (Feature clustering). 

\begin{table*}[!ht]
\centering
\small
    \begin{tabular}{ccccc}
    \hline
	Clustering feature & Metrics & Feature clustering & Siamese + 3 FC & Triplet \\\hline
	\multirow{3}{*}{WELLNAME} &ARI &0.291 $\pm$ 0.071 &\ul{0.508 $\pm$ 0.030} &\textbf{0.635 $\pm$ 0.199} \\
	&AMI &0.451 $\pm$ 0.043 &\ul{0.697 $\pm$ 0.026} &\textbf{0.715 $\pm$ 0.150} \\
	&V-measure &0.452 $\pm$ 0.042 &\ul{0.698 $\pm$ 0.026} &\textbf{0.715 $\pm$ 0.150} \\\hline
	\multirow{3}{*}{CLASS} &ARI &0.340 $\pm$ 0.087 &\textbf{0.655 $\pm$ 0.171} &\ul{0.571 $\pm$ 0.208} \\
	&AMI &0.447 $\pm$ 0.107 &\textbf{0.726 $\pm$ 0.099} &\ul{0.666 $\pm$ 0.160} \\
	&V-measure &0.447 $\pm$ 0.107 &\textbf{0.727 $\pm$ 0.099} &\ul{0.667 $\pm$ 0.160} \\\hline
	\multirow{3}{*}{CLASS+LAYER} &ARI &0.340 $\pm$ 0.087 &\textbf{0.655 $\pm$ 0.171} &0.325 $\pm$ 0.065 \\
	&AMI &0.447 $\pm$ 0.107 &\textbf{0.726 $\pm$ 0.099} &0.494 $\pm$ 0.045 \\
	&V-measure &0.447 $\pm$ 0.107 &\textbf{0.727 $\pm$ 0.099} &0.497 $\pm$ 0.044 \\
	\hline
    \end{tabular}
    \caption{Comparison of the quality of models for Linking problem}
    \label{tab:well_linking_clustering}
\end{table*}

\begin{table*}[!ht]
\centering
\small
    \begin{tabular}{ccccc}
    \hline
	Clustering feature & Metrics & Feature clustering &  Siamese + 3 FC & Triplet \\ \hline
	\multirow{3}{*}{WELLNAME} &ARI &0.291 $\pm$ 0.071 &\ul{0.532 $\pm$ 0.078} &\textbf{0.663 $\pm$ 0.139} \\
	&AMI &0.451 $\pm$ 0.043 &\ul{0.656 $\pm$ 0.053} &\textbf{0.753 $\pm$ 0.097} \\
	&V-measure &0.452 $\pm$ 0.042 &\ul{0.656 $\pm$ 0.053} &\textbf{0.754 $\pm$ 0.096} \\\hline
	\multirow{3}{*}{CLASS} &ARI &0.340 $\pm$ 0.087 &\ul{0.416 $\pm$ 0.147} &\textbf{0.618 $\pm$ 0.225} \\
	&AMI &0.447 $\pm$ 0.107 &\ul{0.557 $\pm$ 0.109} &\textbf{0.708 $\pm$ 0.146} \\
	&V-measure &0.447 $\pm$ 0.107 &\ul{0.557 $\pm$ 0.109} &\textbf{0.708 $\pm$ 0.146} \\\hline
	\multirow{3}{*}{CLASS+LAYER} &ARI &0.340 $\pm$ 0.087 &\textbf{0.416 $\pm$ 0.147} &\ul{0.361 $\pm$ 0.064} \\
	&AMI &0.447 $\pm$ 0.107 &\textbf{0.557 $\pm$ 0.109} &\ul{0.502 $\pm$ 0.065} \\
	&V-measure &0.447 $\pm$ 0.107 &\textbf{0.557 $\pm$ 0.109} &\ul{0.505 $\pm$ 0.064} \\
	
	\hline
    \end{tabular}
    \caption{Comparison of the clustering quality of models for Close Linking problem}
    \label{tab:close_well_linking_clustering}
\end{table*}

As can be seen, the neural networks trained in unsupervised manners strongly correlate with the expert's labelling, significantly improved results from original features. In other words, we provide evidence that the obtained representations are arranged in space according to geological nature, i.e. the similar objects lie closer to each other and are divided from dissimilar. 

\subsection{Sanity and Robustness analysis}\label{sec:sanity_checks}

To  analyse  how  much  data  we  need  to  obtain  a good  data-based  model,  we  conduct the following experiments:
\begin{itemize}
\item We change the number of samples used for training of models from $100$ to $25000$ objects. Thus, we restrict the available amount of the data.
\item We change the number of wells used for sampling $10000$ objects in the training set. Thus, we increase the diversity of objects between samples.
\end{itemize}

\begin{figure*}[!ht]
    \centering
    \includegraphics[width=\linewidth]{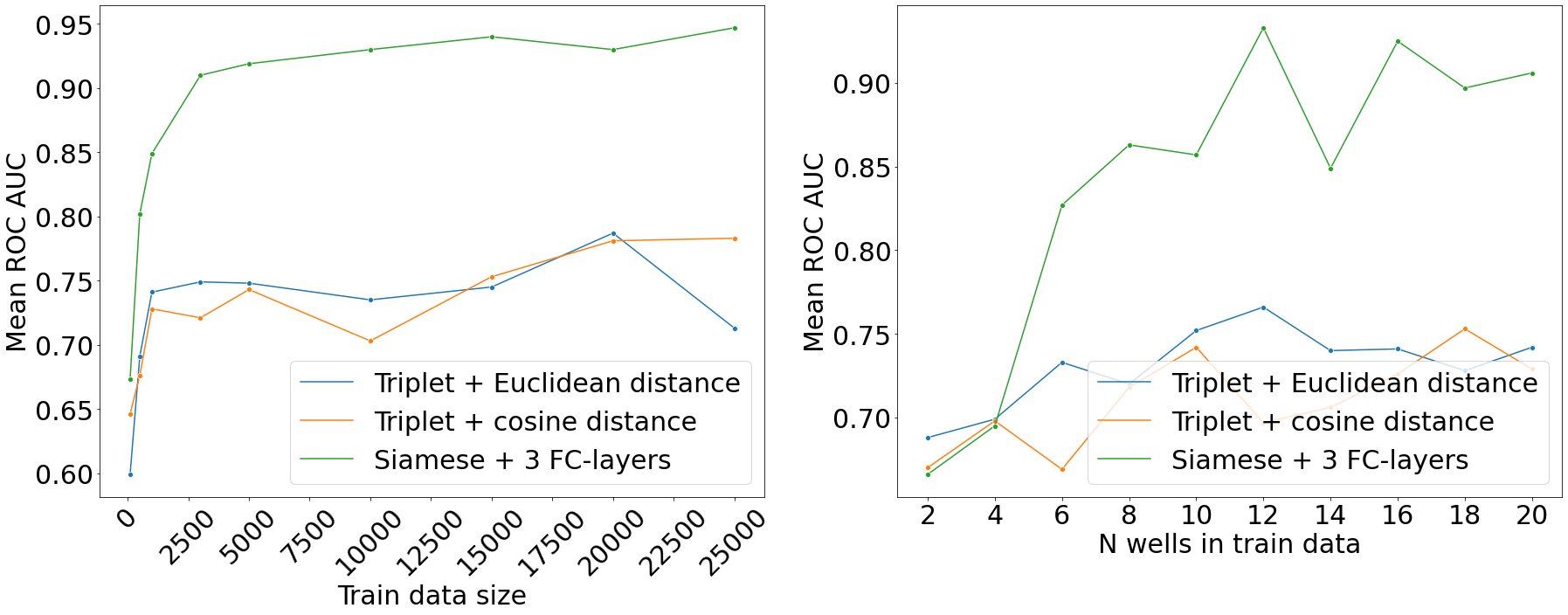}
    \caption{The dependence of models' ROC AUC scores on the \emph{train data size} (left) and the \emph{number of wells} in train data (right) for Linking problem}
    \label{fig:roc_auc_train_size_n_wells_well_linking}
\end{figure*}

\begin{figure*}[!ht]
    \centering
    \includegraphics[width=\linewidth]{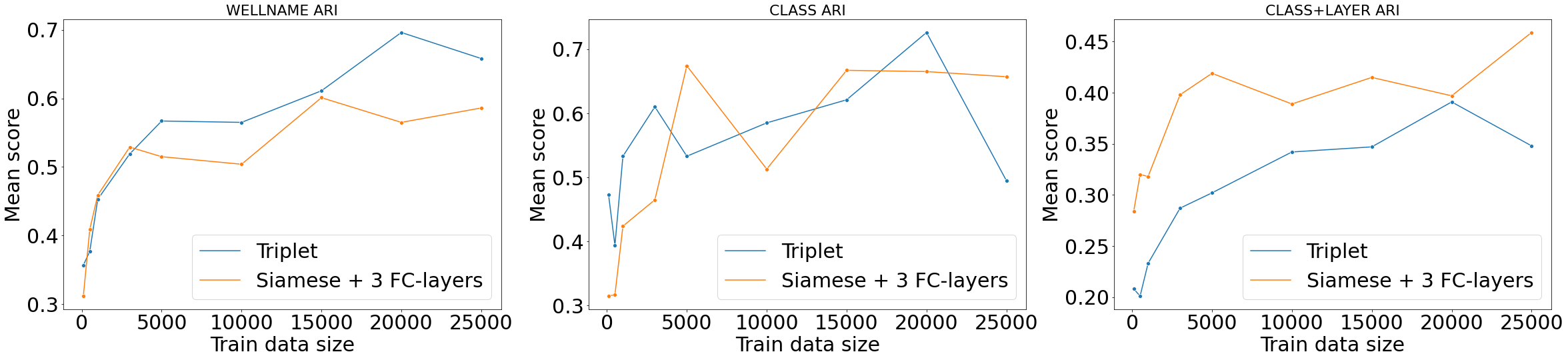}
    \caption{The dependence of models' ARI on the training data size for Linking problem. We consider three possible true labellings from left to right: WELLNAME, CLASS, CLASS+WELLNAME.}
    \label{fig:ari_clustering_train_size_well_linkng}
\end{figure*}

\begin{figure*}[!ht]
    \centering
    \includegraphics[width=\linewidth]{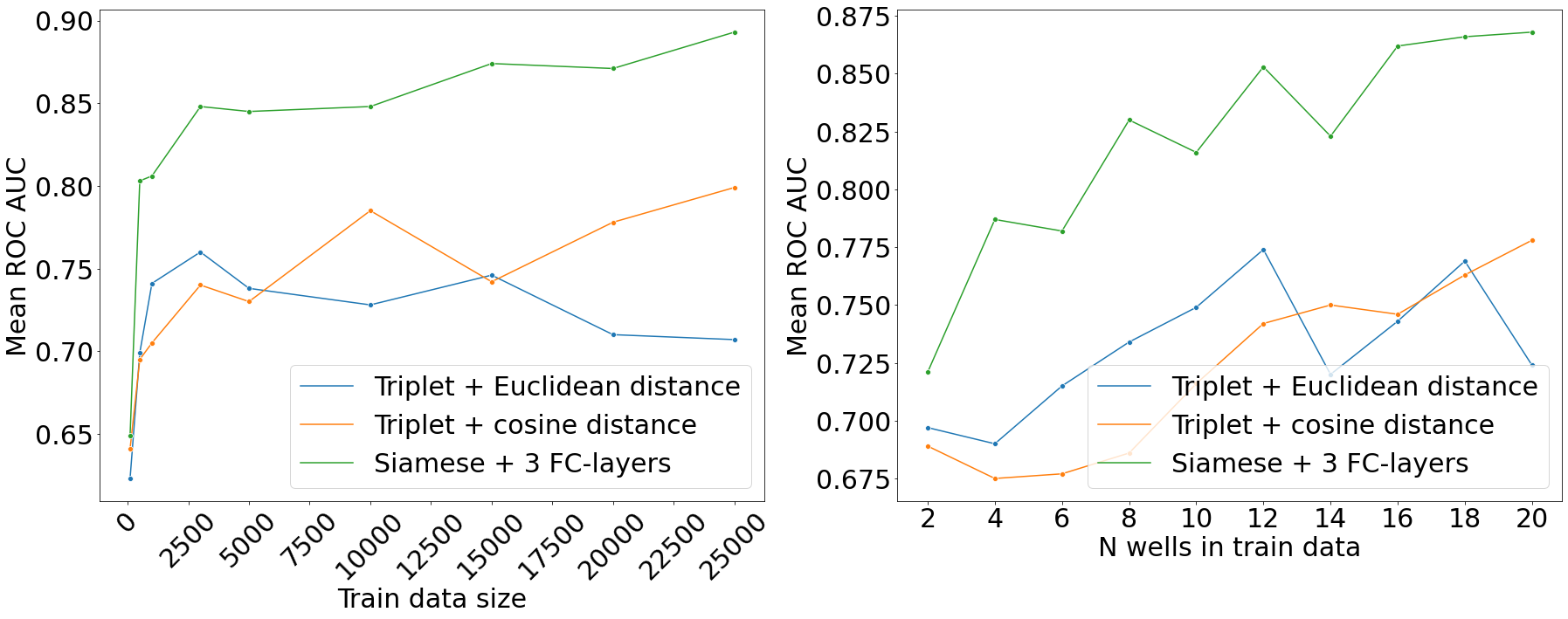}
    \caption{The dependence of models' ROC AUC scores on the \emph{train data size} (left) and the \emph{number of wells} in train data (right) for Close linking problem}
    \label{fig:roc_auc_train_size_n_wells_close_well_linking}
\end{figure*}

\begin{figure*}[!ht]
    \centering
    \includegraphics[width=\linewidth]{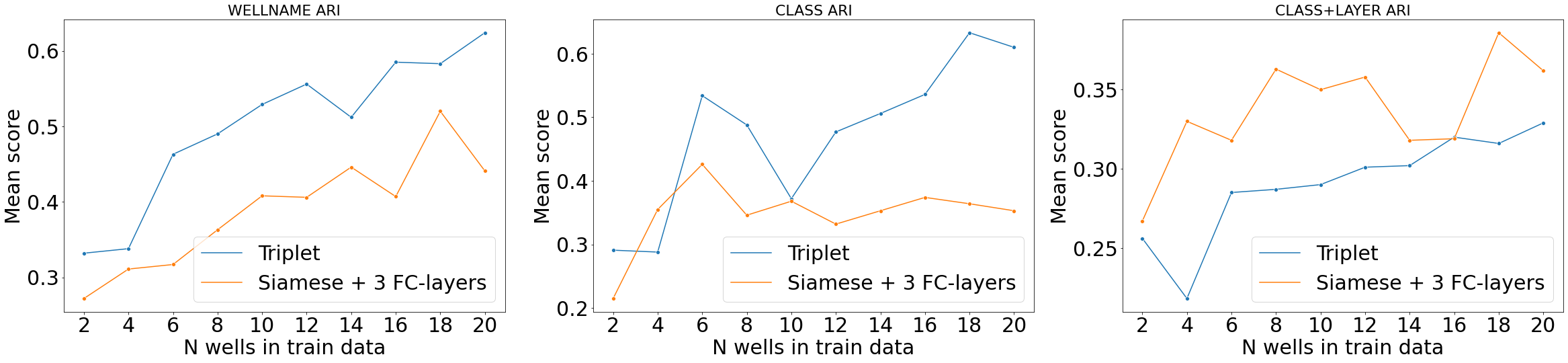}
    \caption{The dependence of models' ARI on the number of wells in train data for Close linking problem}
    \label{fig:ari_clustering_train_n_wells_well_close_linkng}
\end{figure*}

The results for ROC AUC metric are presented in Figures \ref{fig:roc_auc_train_size_n_wells_well_linking} and \ref{fig:roc_auc_train_size_n_wells_close_well_linking} and the results for ARI metric are shown in Figures \ref{fig:ari_clustering_train_size_well_linkng} and \ref{fig:ari_clustering_train_n_wells_well_close_linkng}. 

We can see that 10000 pairs of slices are enough for the model’s training, and further increase of the training data size provides little improvement of the model’s quality. At the same time, the greater number of wells in the training data increases the model’s quality.

\begin{figure*}[!ht]
    \centering
    \includegraphics[width=\linewidth]{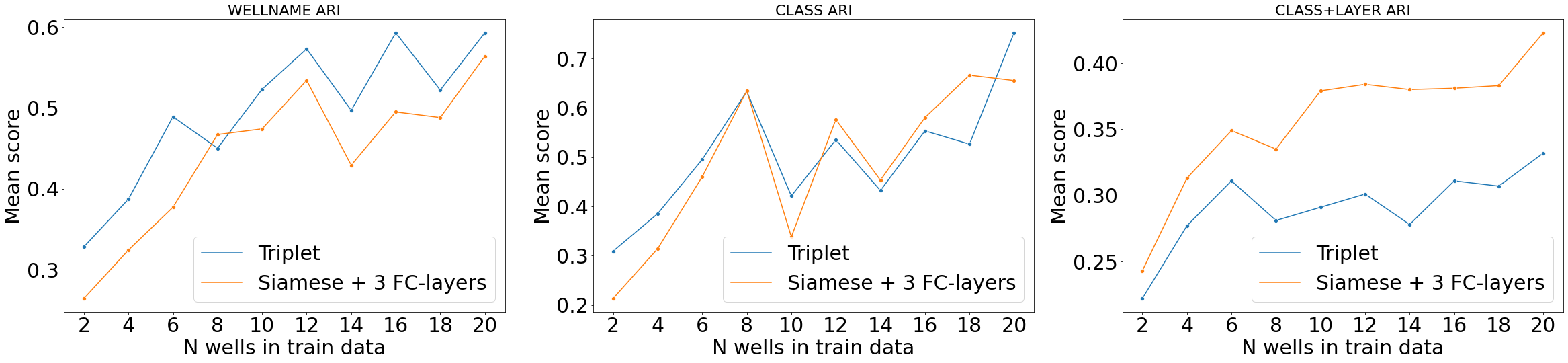}
    \caption{The dependence of models' ARI on the number of wells in train data for Linking problem}
    \label{fig:ari_clustering_train_n_wells_well_linkng}
\end{figure*}

\begin{figure*}[!ht]
    \centering
    \includegraphics[width=\linewidth]{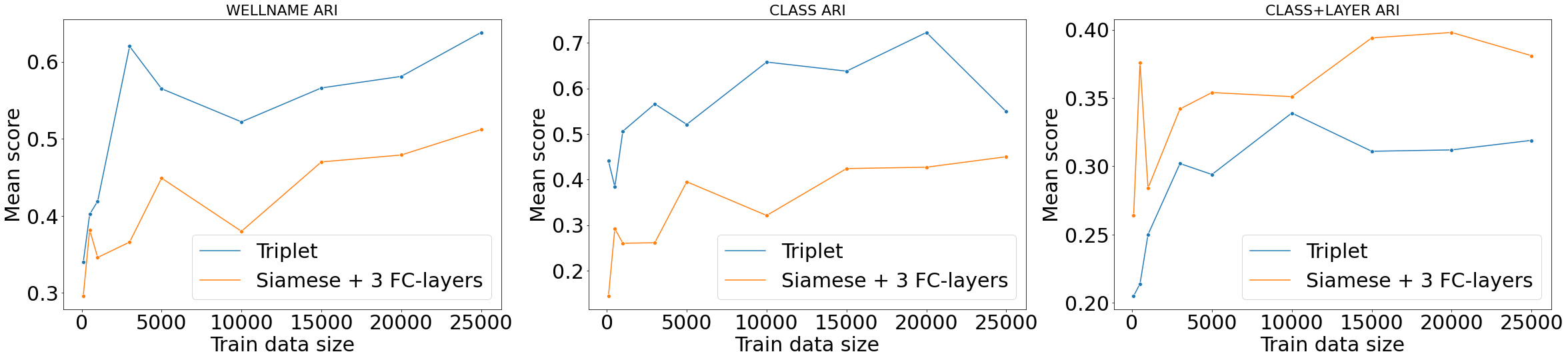}
    \caption{The dependence of models' ARI on the train data size for CLOSE WELL LINKING problem}
    \label{fig:ari_clustering_train_size_close_well_linkng}
\end{figure*}

\begin{figure*}[!ht]
    \centering
    \includegraphics[width=0.8\linewidth]{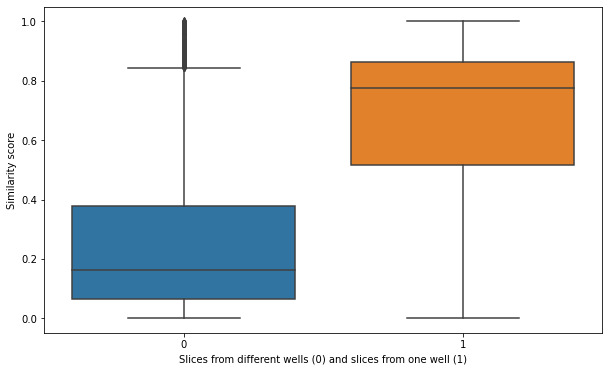}
    \caption{Box plot of models' similarity scores for intervals from different wells and similar wells. Horizontal lines correspond to the mean similarity in the training sample.}
    \label{fig:box_plot_sanity}
\end{figure*}

To assess the adequacy of the models, we also evaluate how well our models distinguish intervals from similar wells and different wells. For this purpose, we generate pairs of intervals from the same wells and the different wells and visualise box plots for similarity scores of each model in Figure~\ref{fig:box_plot_sanity}. 
As can be noticed, most of the obtained scores for similar intervals are close to 1, and the obtained scores for different intervals are close to zero, as expected. Only Logistic regression stands out, having similar results for both categories, proving one more time that this model performs badly. 

\begin{figure*}[!ht]
    \centering
    \includegraphics[width=0.7\linewidth]{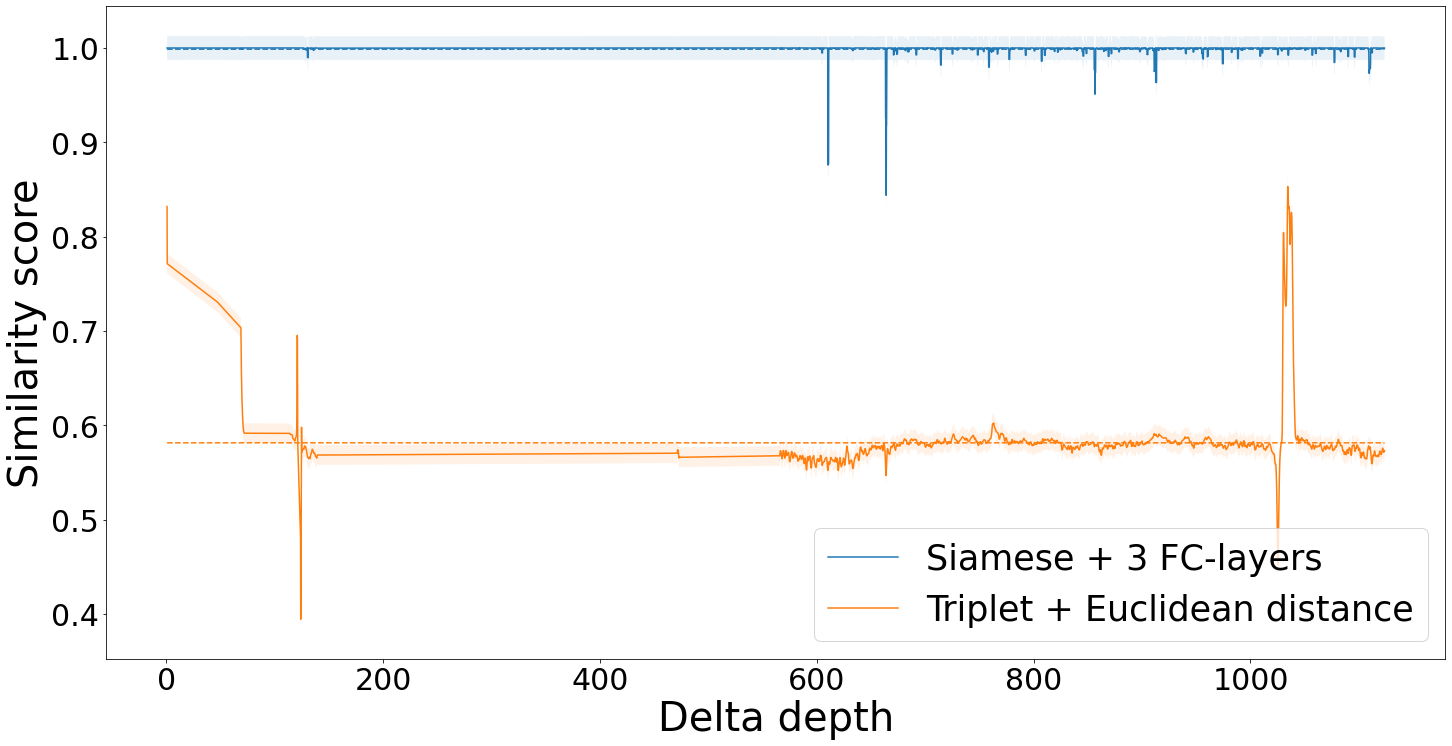}
    \caption{The dependence of models trained on Linking problem similarity scores on the depth difference. Horizontal lines correspond to the mean similarity in the training sample.}
    \label{fig:sim_scores_delta}
\end{figure*}

\begin{figure*}[!ht]
    \centering
    \includegraphics[width=0.7\linewidth]{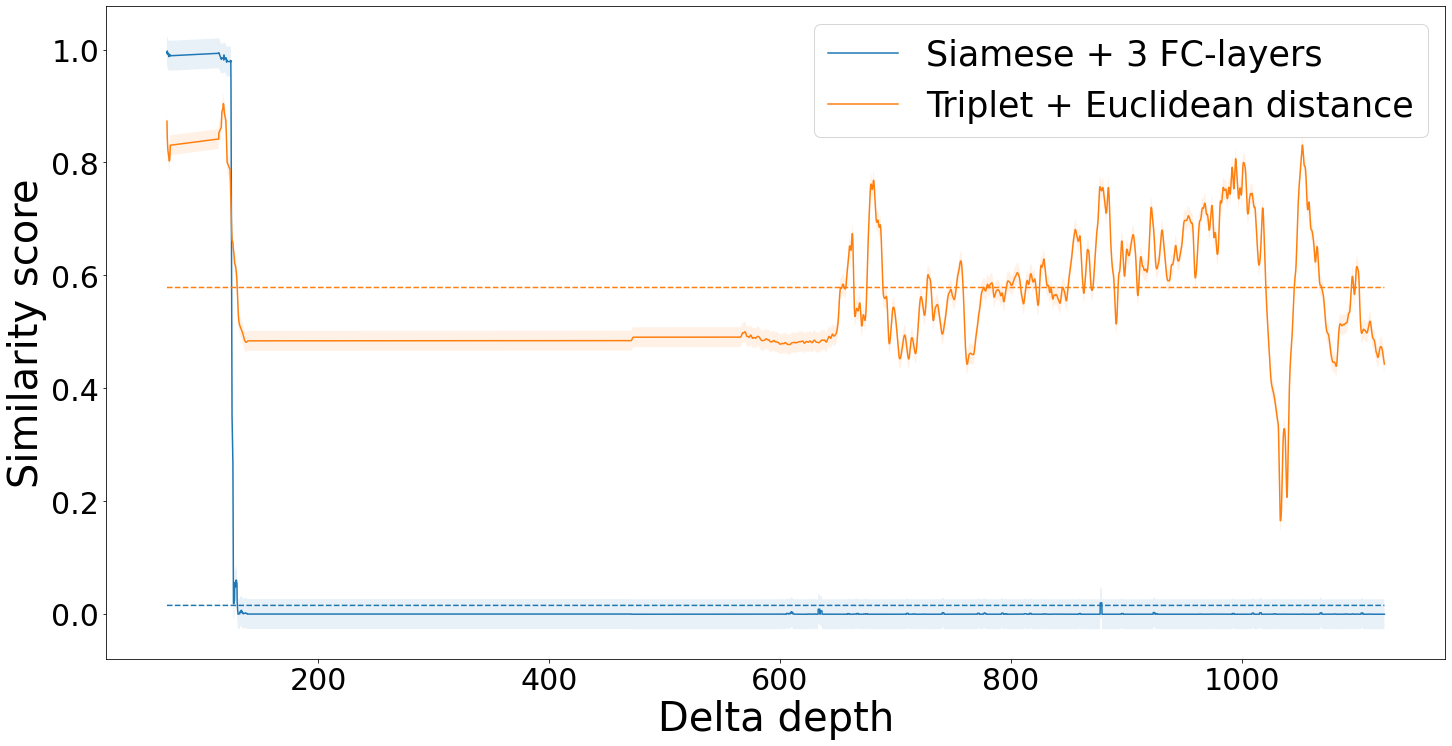}
    \caption{The dependence of models trained on Close linking problem similarity scores on the depth difference}
    \label{fig:sim_scores_delta_close}
\end{figure*}

In addition, we analyse the dependency of similarity score between intervals pair with the delta between their depths for best neural models. We consider one particular well picked for its long length. We shift the starting depth for the second interval from it for this experiment. The result resented in Figure \ref{fig:sim_scores_delta} shows that for Triplet models, the bigger the depth difference, the lower the similarity score, while for the Siamese approach, there is no evident dependency at the presented scale. So, our Triplet models meet natural expectations about changing a well's properties with an increase in depth.
For the Siamese approach, the model shows the same similarity close to $1$ along the well. 
This behaviour is also good, as it is decoded on our loss function.
Moreover, if we consider another loss function Figure~\ref{fig:sim_scores_delta_close} provides desired behaviour for both Siamese and Triplet models.

\subsection{Basic aggregation strategies}\label{sec:naive_agg}

Our neural models provide embeddings and compare only intervals due to the technical limitation for training neural network models. However, the target task is to compare the whole wells. So, we need some aggregation techniques. 

The basic approach is to apply pooling by averaging embedding along the well or taking maximum or minimum values. To compare the basic strategies, we consider the following methods for aggregating the set of intervals' embedding for every well:
\begin{itemize}
\item The result well's embedding/similarity score is a \underline{mean} of corresponded intervals' embeddings/scores;
\item The result well's embedding/similarity is a \underline{maximum} (or \underline{minimum}) between corresponded intervals' embeddings/scores;
\item The result well's embedding/similarity is a \underline{standard} \underline{deviation} of corresponded intervals' embed\-dings/sco\-res.
\end{itemize}

At the same time, when we aggregate obtained embedding, we call it \emph{micro} aggregation. In contrast, when we aggregate obtained similarity scores, we denote it as \emph{macro} aggregation. 

The results for both types of aggregation are averaged by 2 folds and presented in Tables~\ref{tab:macro_agg} and \ref{tab:micro_agg}. 

\begin{table*}[!ht]
\centering
\begin{tabular}{c|ccc}
\hline
\multicolumn{4}{c}{\textbf{Triplet model}} \\
\hline
Aggregation strategy & ARI & AMI & V\_measure \\
\hline
max & 0.153 $\pm$ 0.204 & 0.188 $\pm$ 0.253 & 0.653 $\pm$ 0.073 \\
min & -0.065 $\pm$ 0.033 & -0.062 $\pm$ 0.071 & 0.507 $\pm$ 0.025 \\
mean & -0.081 $\pm$ 0.033 & -0.135 $\pm$ 0.097 & 0.453 $\pm$ 0.112 \\
std & \textcolor{blue}{\textbf{0.354 $\pm$ 0.233}} & \textcolor{blue}{\textbf{0.398 $\pm$ 0.202}} & \textcolor{blue}{\textbf{0.738 $\pm$ 0.061}} \\
\hline
\multicolumn{4}{c}{\textbf{Siamese + 3FC model}} \\
\hline
& ARI & AMI & V\_measure \\
\hline
max & 0.096 $\pm$ 0.025 & 0.136 $\pm$ 0.059 & 0.607 $\pm$ 0.071 \\
min & -0.169 $\pm$ 0.044 & -0.247 $\pm$ 0.070 & 0.400 $\pm$ 0.107 \\
mean & 0.058 $\pm$ 0.109 & 0.055 $\pm$ 0.120 & 0.578 $\pm$ 0.002 \\
std & \textbf{0.182 $\pm$ 0.022} & \textbf{0.250 $\pm$ 0.018} & \textbf{0.671 $\pm$ 0.028} \\
\hline
\end{tabular} 
\caption{Comparison of \emph{macro} aggregation strategies, i.e. aggregation of scores, for various models estimated for CLASS expert's labelling. Best values for each model are highlighted with \textbf{bold} font and overall best values are \textcolor{blue}{\textbf{blue}}.}
    \label{tab:macro_agg}
\end{table*}

\begin{table*}[!ht]
\centering
\begin{tabular}{c|ccc}
\hline
\multicolumn{4}{c}{\textbf{Triplet model}} \\
\hline
Aggregation strategy & ARI & AMI & V\_measure \\
\hline
max & 0.281 $\pm$ 0.210 & 0.352 $\pm$ 0.227 & 0.691 $\pm$ 0.140 \\
min & 0.327 $\pm$ 0.007 & 0.425 $\pm$ 0.038 & 0.730 $\pm$ 0.040 \\
mean & \textbf{0.432 $\pm$ 0.197} & \textbf{0.444 $\pm$ 0.180} & \textbf{0.758 $\pm$ 0.051} \\
std & 0.174 $\pm$ 0.112 & 0.186 $\pm$ 0.094 & 0.641 $\pm$ 0.003 \\
\hline
\multicolumn{4}{c}{\textbf{Siamese + 3FC model}} \\
\hline
& ARI & AMI & V\_measure \\
\hline
max & 0.171 $\pm$ 0.092 & 0.233 $\pm$ 0.122 & 0.667 $\pm$ 0.009 \\
min & 0.282 $\pm$ 0.025 & 0.339 $\pm$ 0.048 & 0.702 $\pm$ 0.052 \\
mean & \textcolor{blue}{\textbf{0.547 $\pm$ 0.055}} & \textcolor{blue}{\textbf{0.540 $\pm$ 0.067}} & \textcolor{blue}{\textbf{0.793 $\pm$ 0.005}} \\
std & 0.179 $\pm$ 0.107 & 0.259 $\pm$ 0.115 & 0.666 $\pm$ 0.020 \\
\hline
\end{tabular} 
\caption{Comparison of \emph{micro} aggregation strategies, i.e. aggregation of embeddings, for various models estimated for CLASS expert's labelling. Best values for each model are highlighted with \textbf{bold} font and overall best values are \textcolor{blue}{\textbf{blue}}.}
    \label{tab:micro_agg}
\end{table*}

As expected, when working at the level of the received scores, i.e. macro aggregation, we lose much valuable information, and, as a result, the obtained metrics are poor. In addition, this approach is very sensitive to the choice of intervals. We hypothesise that it is necessary to cut each well into all possible intervals and compare all pairs to achieve good results. Such a procedure will take considerable time. 

In contrast, macro aggregation seems to be a better strategy. We see that embedding aggregation by mean provides the best results with respect to all clustering metrics. It means that the behaviour of obtained embeddings on a well-level is consistent with the expert's expectation about them. 

\subsection{Aggregation via end-to-end model}\label{sec:end2end_results}

End-to-end approach uses expert labelling to train the aggregation model, so it can perform better compared to a pure self-supervised scenario. For evaluation of the performance of the end-to-end approach, we analyse three different techniques of the model's training:
\begin{enumerate}
    \item \underline{Pure end-to-end:} training both embedding part and aggregation part from scratch;
    \item \underline{Only aggregation:} pre-trained embedding part is fixed, only parameters of aggregation model change;
    \item \underline{From checkpoint:} we use a pre-trained embedding part, but we fine-tune it simultaneously with the training of the aggregation model from scratch;
\end{enumerate}

We utilize the Triplet model for the embedding part, as it seems to provide the universal embeddings. As an architecture of the aggregation model, we use the LSTM layer with hidden size $32$ followed by ReLU activation, output Linear layer, Dropout with probability $0.25$ and Softmax as activation function. We accurately hold out test wells to have an example of each CLASS in both train and test datasets. We train a model for the Pure end-to-end approach for 17 epochs, for Only aggregation - 6 epochs and for From checkpoint method, we use 20 epochs, as such selection of hyperparameters allows to train for a reasonable time and avoid overfitting. The results are averaged for two test datasets.

Obtained results are in Table~\ref{tab:end2end}.
The end-to-end model trained in a proper way performs better than the basic aggregation strategy, making it a better solution, as expected. We hypothesise that increasing the numbers of wells for the training dataset may lead to an even better result. 
At the moment, Only end-to-end aggregation strategy demonstrates the most decent results, showing that our embeddings are good as they are.

\begin{table*}[h!]
    \centering
    \begin{tabular}{c|ccc}
    \hline
    & ARI & AMI & V\_measure \\
    \hline
    Pure end-to-end & 0.476 $\pm$ 0.083 & 0.591 $\pm$ 0.137 & 0.807 $\pm$ 0.010 \\
    Only aggregation & \textbf{0.629 $\pm$ 0.087} & \textbf{0.720 $\pm$ 0.025} & \textbf{0.845 $\pm$ 0.071} \\
    From checkpoint & 0.309 $\pm$ 0.085 & 0.444 $\pm$ 0.011 & 0.697 $\pm$ 0.120 \\
    \hline
    \end{tabular}     
    \caption{Metrics for end-to-end approach for three training strategies}
    \label{tab:end2end}
\end{table*}

\subsection{Domain adaptation}\label{sec:transfer_learning_results}

Our deep learning model should have a high generalization property and adapt to other domains. After training with diverse enough data, we should be able to apply it to the data from a different formation or a different field. 
The best scenario is the ability to use the model as it is.
But it is also a good solution when we slightly fine-tune our model using a small amount of new data.

To evaluate our approach in domain adaptation task, we consider a set of formations and fields: New Zealand's formations Urenui, Marinui, Matemateaonga, Mohakatino, Moki and principled different Norway's formation Utsira.

During experiments, we evaluate three different methods to work with a new domain: 
\begin{itemize}
    \item \emph{Pre-trained:} straightforward applying the model pre-trained on one domain to the new data;
    \item \emph{From scratch:} use the newly trained model without applying an old model;
    \item \emph{Fine-tuning:} started from weights of pre-trained model, fine-tune the model to a new data. 
    
\end{itemize}

For comparison, we consider the Triplet model with Euclidean distance trained for Linking problems. The results are presented in Table~\ref{tab:ari_tranfser_class_layer_triplet}.

\begin{table*}[!ht]
\centering
\small
\begin{tabular}{p{2.5cm}|ccc|ccc}
\hline
& Pre-tr. & From scratch & Fine-tuning & Pre-tr. & From scratch & Fine-tuning \\
\hline
Metrics & \multicolumn{3}{c}{Marinui} & \multicolumn{3}{c}{Matemateaonga} \\
\hline
\footnotesize{ROC AUC} & 0.693 & 0.780 $\pm$ 0.011 & 0.761 $\pm$ 0.020 & 0.695 & 0.755 $\pm$ 0.009 & 0.772 $\pm$ 0.005 \\
\footnotesize{ARI WELL.} & 0.24 & 0.437 $\pm$ 0.034 & 0.489 $\pm$ 0.025 & 0.237 & 0.416 $\pm$ 0.031 & 0.515 $\pm$ 0.019 \\
\footnotesize{ARI CLASS}  & 0.183 & 0.278 $\pm$ 0.023 & 0.282 $\pm$ 0.014 & 0.109 & 0.161 $\pm$ 0.016 & 0.203 $\pm$ 0.036 \\
\footnotesize{ARI CL.+ LAY.} & 0.143 & 0.235 $\pm$ 0.015 & 0.247 $\pm$ 0.011 & 0.125 & 0.172 $\pm$ 0.017 & 0.195 $\pm$ 0.010\\
\hline
& \multicolumn{3}{c}{Mohakatino} & \multicolumn{3}{c}{Moki} \\
\hline
\footnotesize{ROC AUC} & 0.730 & 0.863 $\pm$ 0.023 & 0.862 $\pm$ 0.020 & 0.651 & 0.815 $\pm$ 0.019 & 0.789 $\pm$ 0.009 \\
\footnotesize{ARI WELL.} & 0.231 & 0.575 $\pm$ 0.099 & 0.676 $\pm$ 0.048 & 0.263 & 0.501 $\pm$ 0.083 & 0.593 $\pm$ 0.050 \\
\footnotesize{ARI CLASS} & 0.287 & 0.294 $\pm$ 0.044 & 0.330 $\pm$ 0.109 & 0.158 & 0.239 $\pm$ 0.018 & 0.273 $\pm$ 0.014 \\
\footnotesize{ARI CL.+ LAY.} & 0.123 & 0.231 $\pm$ 0.034 & 0.255 $\pm$ 0.019 & 0.122 & 0.222 $\pm$ 0.022 & 0.237 $\pm$ 0.020 \\
\hline
\end{tabular} 
    \caption{ROC AUC and ARI (trained on Linking problems and clustered on intervals) for domain adaptation of Triplet model. We consider three expert labellings: WELLNAME labelling, CLASS labelling and CLASS+LAYER labelling (CL.+LAY.)}
    \label{tab:ari_tranfser_class_layer_triplet}
\end{table*}

We see that the model pre-trained on Urenui formation demonstrates tolerable quality for all formations compared with models trained from scratch. Nevertheless, using pre-trained weights with further optimizing it to new data shows the best results in most cases, significantly improving model quality.

Our approach shows similar behaviour even while applying for the formation with absolutely different geological characteristics—the results for Norway formation presented in Table~\ref{tab:ari_tranfser_class_layer_triplet_norway}.  

\begin{table*}[!ht]
\centering
\small
\begin{tabular}{p{2.5cm}|cccccc}
\hline
Metrics &Pre-tr. & From scratch &Fine-tuning \\ \hline
\footnotesize{ROC AUC} &0.765 &0.836 $\pm$ 0.029 &0.803 $\pm$ 0.038 \\
\footnotesize{ARI WELL.} &0.469 &0.699 $\pm$ 0.006 &0.560 $\pm$ 0.091 \\
\footnotesize{ARI CLASS} &0.709 &0.558 $\pm$ 0.050 &0.660 $\pm$ 0.049 \\
\footnotesize{ARI CL.+ LAY.} &0.65 &0.755 $\pm$ 0.061 &0.656 $\pm$ 0.006 \\
\hline
\end{tabular} 
    \caption{ROC AUC and ARI (trained on Linking problems and clustered on intervals) for domain adaptation of Triplet model for Norway dataset. We consider three expert labellings: for WELLNAME labelling, CLASS labelling and CLASS+LAYER labelling (CL.+LAY.)}
    \label{tab:ari_tranfser_class_layer_triplet_norway}
\end{table*}

The results obtained indicate an excellent ability of the model to work on new formations, which is essential for further application of the model in the oil\&gas industry.

\section{Conclusions}\label{sec:conclusions}

We managed to construct a data-based model that evaluates a similarity between wells. The model similarity reflects the closeness of petrophysical and geological properties between wells and uses logging data as input. 
The model is a deep neural network for processing sequential data. The key innovation is the usage of the triplet loss function for the model and specific architecture selected among a wide family of possible architectures.

The choices we made lead to the two key beneficial properties of our model are (1) self-supervised training regime and (2) meaningful (from a geological point of view) representations of wells and well intervals. 

Further analysis via the usage of obtained representations for clustering wells reveals that they reflect expert-provided labelling of wells. 

For NN with triplet loss function, we see that it outperforms baselines based on classic machine learning approaches, as quality metrics widely used in literature for the problem at hand for our model are better. 
In particular, we want a model to maximize two main metrics for classification problem area under ROC curve ROC AUC and area under Precision-Recall curve PR AUC.
When comparing intervals, we get quality measures ROC AUC, PR AUC, and Accuracy values are as high as $0.783$, $0.926$, and $0.926$ correspondingly for our model cf. $0.802$, $0.895$, and $0.787$ for a baseline gradient boosting model. 
We also consider the quality of similarity metrics via Adjusted Rand Index (ARI) and Adjusted Mutual Information (AMI), which we also want to maximize.
When comparing wells, we get quality measure ARI and AMI as high as $0.635$ and $0.715$ compared to $0.291$ and $0.451$ for the baseline.

In addition to basic experiments, we considered transfer learning and end-to-end learning. In the transfer learning scenario, our models also perform well. Thus, it is possible to use constructed models for similarity identification on new formations. We obtain ARI $0.469$ comparable to ARI value $0.699$ obtained via training a similarity model using new data for a different Norway data. Our results suggest that via deep learning, we can construct a decent end-to-end model with ARI $0.629$ via a self-supervised approach and a simple model ahead of it. Thus, we can achieve high quality with smaller labelled data size. 

We expect that bigger samples of wells can provide even more interesting results, especially if the data for model training are diverse enough.
Detailed research in this direction can lead to new insights on how similarity models and representations based on them can help in various oil\&gas applied problems.

\printcredits

\section*{Acknowledgements}
We are grateful to Liliya Mironova for assistance with code formatting. We thank Evgeny Burnaev for providing computational resources. 

\bibliographystyle{cas-model2-names}

\bibliography{cas-refs.bib}



\end{document}